\documentclass[journal=jacsat,manuscript=article]{achemso}

\usepackage[version=3]{mhchem} 
\usepackage[htt]{hyphenat}
\usepackage{subcaption}
\usepackage{multirow}

\author{Peter Pak}
\affiliation{
  Department of Mechanical Engineering, Carnegie Mellon University, Pittsburgh,
  PA, USA
}
\alsoaffiliation{Tripoli Rocketry Association, Pittsburgh, PA, USA}

\author{Jesse Barkley}
\affiliation{
  Department of Mechanical Engineering, Carnegie Mellon University, Pittsburgh,
  PA, USA
}
\alsoaffiliation{Tripoli Rocketry Association, Pittsburgh, PA, USA}

\author{Rumi Loghmani}
\affiliation{
  Department of Mechanical Engineering, Carnegie Mellon University, Pittsburgh,
  PA, USA
}
\alsoaffiliation{Tripoli Rocketry Association, Pittsburgh, PA, USA}

\author{Derek Baich}
\affiliation{Tripoli Rocketry Association, Pittsburgh, PA, USA}

\author{Ananya Pamal}
\affiliation{
  Department of Mechanical Engineering, Carnegie Mellon University, Pittsburgh,
  PA, USA
}

\author{Amir Barati Farimani}
\email{barati@cmu.edu}
\affiliation{
  Department of Mechanical Engineering, Carnegie Mellon University, Pittsburgh,
  PA, USA
}
\alsoaffiliation{
  Machine Learning Department, Carnegie Mellon University, Pittsburgh, PA, USA
}

\title[]{
    Rocketsmith: Agentic Additive Manufacturing of High-Powered Rockets 
}

\abbreviations{IR,NMR,UV}

\SectionNumbersOn
\begin{document}

\begin{abstract}
RocketSmith is an agentic system which intelligently automates the DFAM process
for the development of high powered rockets suitable for launch. The system
utilizes a large language model to orchestrate the execution of software tools
to validate design characteristics such as flight stability and generate the
parametric design components for the rocket assembly. A collection of subagents
and skills enable optimization workflows of flight parameters via iteration in
both zero-shot and human-in-the-loop workflows. With this system, four distinct
high power rockets with various motor and assembly configurations were developed
utilizing the unique design capabilities of additive manufacturing. These
assembly components were fabricated using various FDM printers, manually
evaluated for flight readiness, and flight tested at a launch event. From these
tests, all rockets achieved a stable launch and two of the four rockets were
successfully recovered in reflyable condition. The altimeter data validated that
the rockets achieved an altitude 80\% of the expected apogee predicted by the
agentic system, establishing consistency between simulation and
experimentation.
\end{abstract}

\section{Introduction}

\begin{figure}
    \centering
    \includegraphics[width=0.9\linewidth]{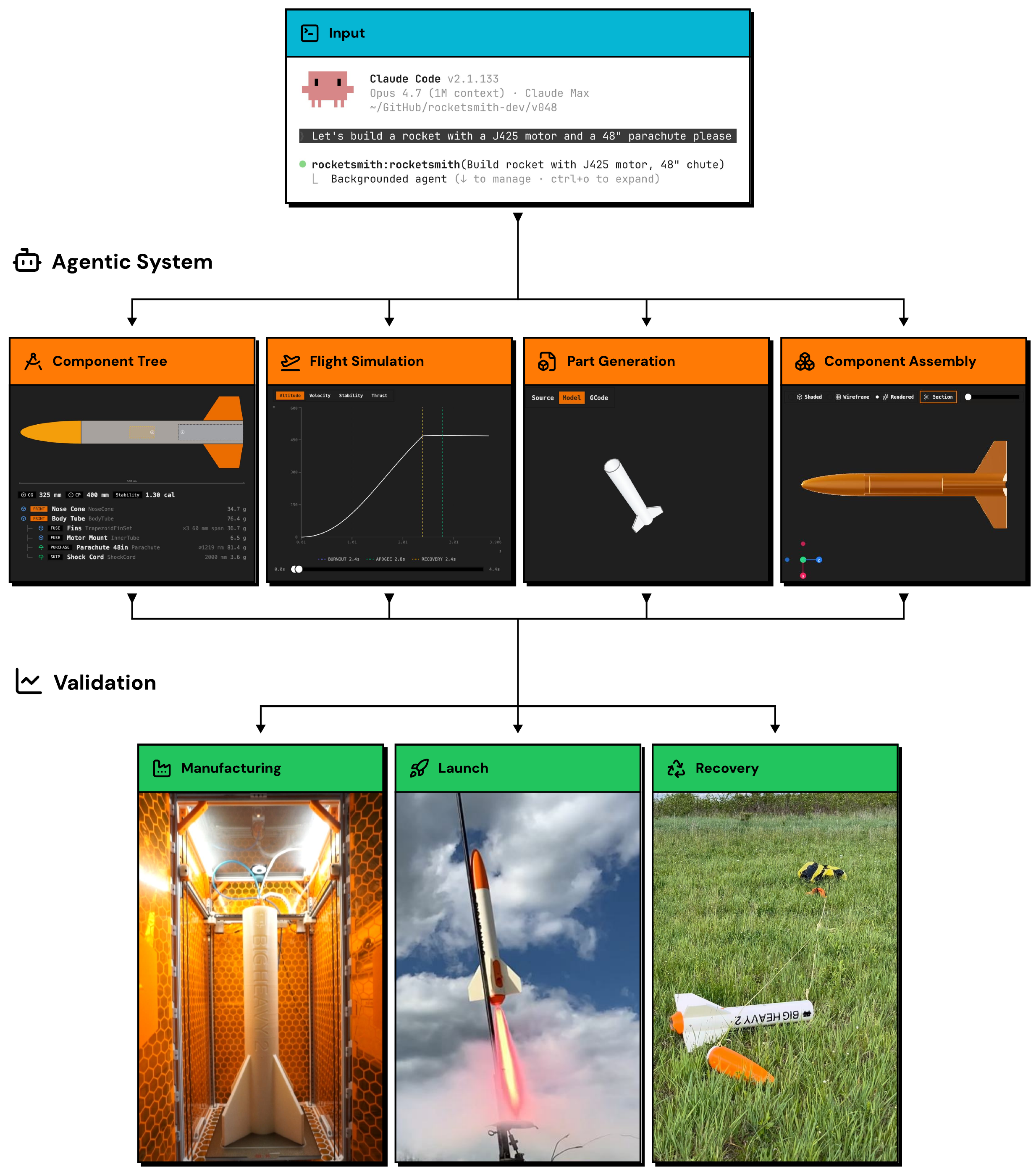}
    \caption{
        RocketSmith utilized as a Claude Code
        \cite{noauthor_anthropicsclaude-code_2026} plugin enables the efficient
        development of high powered rockets. The agentic system is capable of
        designing an OpenRocket \cite{niskanen_openrocket_2015} based component
        tree with provided user constraints such as specific motors and flight
        characteristics. Prescribed dimensions from the component tree are
        utilized to generate parametric models of the rocket airframe using
        CADSmith \cite{barkley_cadsmith_2026}. More precise weight calculations
        obtained with PrusaSlicer \cite{noauthor_prusa3dprusaslicer_2026}
        improve center of gravity estimations and allow for iterative
        optimization of the rocket's stability before manufacturing. Each part
        is additively manufactured via FDM where afterwards the components are
        assembled, flight tested, and successfully recovered.
    }
    \label{fig:main}
\end{figure}

Additive Manufacturing (AM) enables design and fabrication capabilities outside
the realm of conventional manufacturing techniques such as iterative design,
rapid prototyping, and part consolidation \cite{yang_additive_2015}. These
considerations make the utilization of AM attractive to development constrained
and cost elastic industries, specifically aerospace
\cite{blakey-milner_metal_2021}. Within the development of an assembly there are
a number of factors outside the scope of manufacturing that have an effect on
its production application. High powered rocketry presents an environment where
these multifaceted design and manufacturing are encountered and where additive
manufacturing presents a unique solution to part consolidation and flight
optimization. This work explores the development of an agentic system,
RocketSmith (Figure \ref{fig:main}), capable of streamlining the design and
manufacturing process for the development and flight testing of Class 2 high
powered rockets \cite{federal_aviation_administration_requirements_2008}.

Development of high powered rockets is often an iterative process as design
specifications can change due to factors realized in manufacturing, part
availability, and other unforeseen events. This contributes to an iterative loop
where variables such as stability are recalculated with component values
obtained after manufacturing, informing design and manufacturing decisions for
the other components within the assembly. Within the iteration loop there is
noticeable friction between the various software tools for flight simulation,
design, and manufacturing \cite{choi_optimal_2009, chiesa_launch_1999}. The
flight timeline of a high powered rocket is composed of a series of events
summarized as: motor ignition, motor burnout, flight apogee, recovery
deployment, and retrieval. The propellant from motor ignition provides the
necessary thrust to lift the rocket upward until the motor burnout event.
\cite{stine_handbook_2004, sutton_rocket_2026}. The momentum following the motor
burnout event continues to carry the rocket upward until it reaches its apogee
where the deployment event occurs \cite{stine_handbook_2004}. At the apogee,
velocity is minimal and for single deployment systems the main parachute
deployment event occurs. However, for dual deployment systems the drogue chute
deployment occurs at the apogee for a controlled descent before the main
parachute deployment event is executed \cite{stine_handbook_2004}. Following
this sequence of events the final recovery event involves the retrieval of the
rocket from its landing site \cite{stine_handbook_2004}.

Agentic system based approaches to problem solving have displayed considerable
capability in technical fields for applications such as molecular design
\cite{ock_adsorb-agent_2026, ock_catalyst_2023}, robotics
\cite{bartsch_llm-craft_2025, merrill_llm-drone_2025}, additive manufacturing
\cite{jadhav_llm-3d_2025, pak_additivellm_2025, pak_agentic_2026}, software
development\cite{han_tdflow_2026, he_llm-based_2025}, materials science
\cite{nigam_polymer-agent_2026, chaudhari_modular_2026}, and mechanical design
\cite{jadhav_large_2026}. The parametric knowledge embedded within the agentic
system's Large Language Model (LLM) provides the model with not only an enhanced
dataset of training data to extrapolate from but also enables complex reasoning
capability necessary for domain specific tasks \cite{ yao_react_2023,
vaswani_attention_2017}. In addition, these LLMs provide the foundational
reasoning and tool calling abilities allowing interactions with the surrounding
environment \cite{pak_agentic_2026, ock_adsorb-agent_2026, jadhav_large_2026}.
These set of capabilities make the use of agentic systems attractive to the
iterative development cycle of high powered rockets.

\begin{figure}
    \centering
    \includegraphics[width=\linewidth]{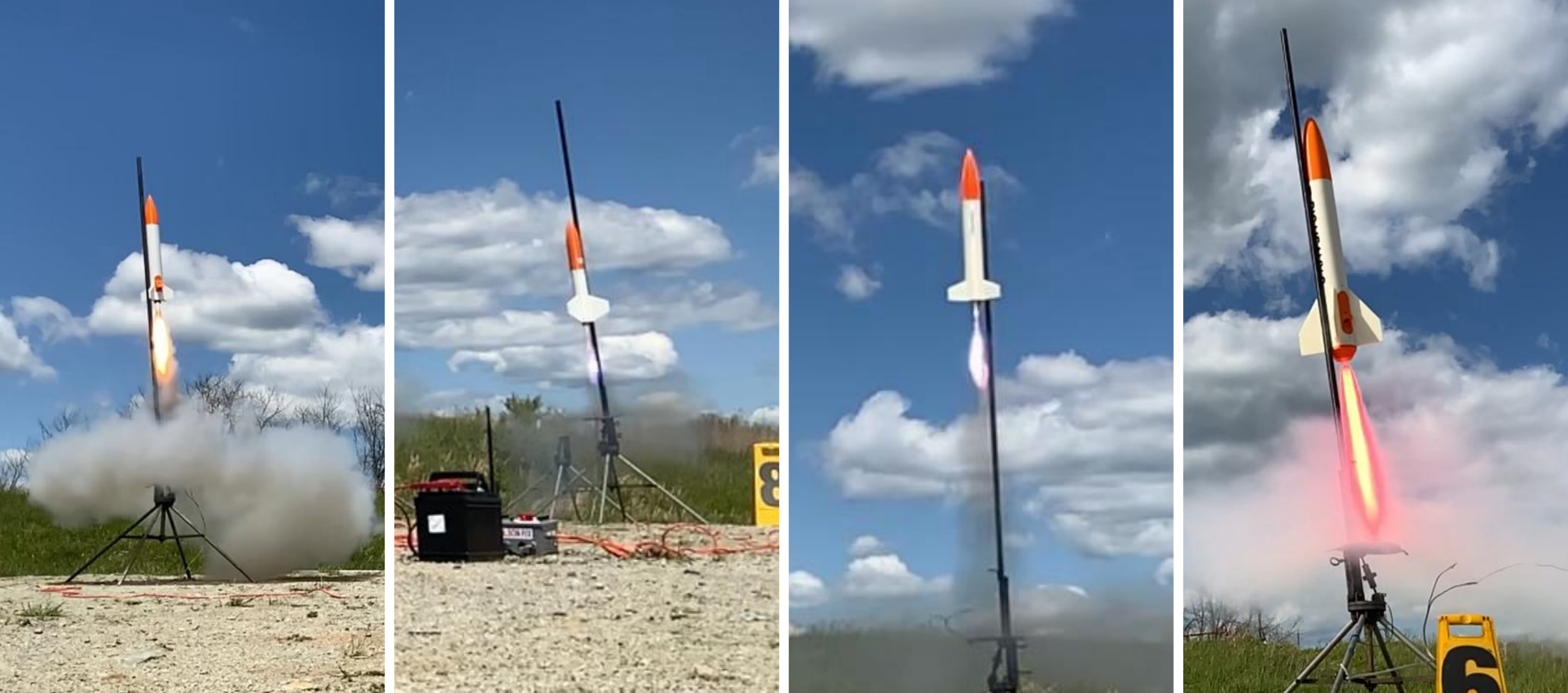}
    \caption{
        (Left to Right) High Power 1 launched by Pak with an AeroTech H100W
        motor, High Power 2 launched by Loghmani with AeroTech H219T motor, High
        Power 3 launched by Barkley also with AeroTech H219T, and High Power 4
        launched by Pak with an AeroTech J425R motor.
    }
    \label{fig:results_launch}
\end{figure}

This work introduces RocketSmith, an agentic system composed of subagents,
skills, and tool calling abilities developed to the standard of the Model
Context Protocol (MCP) and released as a plugin for agent harnesses such as
Claude Code. This system is capable of orchestrating the necessary flight
simulations, designing the appropriate CAD files, and generating manufacturing
files of a high powered rocket with user provided constraints and
specifications. The schematics produced by RocketSmith are then additively
manufactured via Fused Deposition Modeling (FDM) and assembled for evaluation
through a series of flight tests at a following launch event (Figure
\ref{fig:results_launch}). Quantitative altimeter data collected from a subset
of flight tests along with qualitative metrics are used to evaluate the general
performance of the agentic system. The code is offered in the form of a plugin /
extension for Claude Code and is available at
\url{https://github.com/ppak10/RocketSmith}.

\section{Related Works}

LLM-3D print by Jadhav et al. \cite{jadhav_llm-3d_2025} investigates the
application of a large language model (variants of ChatGPT 4 by OpenAI) enabled
agentic system for optimizing fused deposition modeling printing parameters. By
leveraging the vision capabilities of ChatGPT 4, optical images taken of the
build layer are evaluated by the LLM for any potential defects and the necessary
actions are executed to address these issues. The agentic system utilizes the
LangChain library as its primary framework where in which the supervisor agent
orchestrates tasks to its subagents for tasks such as planning, information
extraction, and solution execution \cite{jadhav_llm-3d_2025}. These actions
significantly improved the print quality of tested parts, addressing potential
defects in near real-time, further validated by compression tests with increased
peak load sustained by LLM-3D print fabricated parts compared to parts build
without the assistance of the agentic system \cite{jadhav_llm-3d_2025}.

RocketBench by Simonds et al.\cite{simonds_llms_2025} is a benchmarking tool
which connects LLMs to high-fidelity trajectory simulation of RocketPy
\cite{ceotto_rocketpy_2021} to evaluate each LLM's capability to design rockets.
Models specify designs through a structured JSON interface and are scored on a
composite reward that combines altitude accuracy, structural integrity,
horizontal drift, cost efficiency, and landing safety. The authors find that
while frontier LLMs demonstrate strong baseline engineering intuition, these
models consistently plateau below human expert performance during iterative
refinement, exposing the fundamental limitation that standard LLMs struggle to
translate simulation feedback into meaningful design updates
\cite{simonds_llms_2025}. These findings motivate several architectural
decisions implemented in RocketSmith, where rather than relying on the LLM to
iteratively optimize design parameters directly, delegating these numerically
precise tasks to tools would result in accurate responses. These include tasks
such as stability calculation, trajectory simulation, slicing, and mass
estimation. This addresses the core failure mode identified in RocketBench where
instead of asking the LLM to reason about simulation feedback and update
parameters itself, the design process is structured so that each stage produces
verifiable outputs for the next stage to consume deterministically.

Early text-to-CAD approaches such as CAD-LLM \cite{sifan_wu_cad-llm_2023} and
LLM4CAD \cite{li_llm4cad_2024} framed CAD generation as a sequence modeling
problem, prompting models to produce CAD command sequences from natural
language. These methods demonstrated basic geometric reasoning but struggled
with dimensional precision, as parametric errors compound through a modeling
sequence without any external correction signal. More recent systems have moved
toward the Agent-Aided Design paradigm, placing an agent in a feedback loop that
iteratively generates, compiles, and visually inspects geometry
\cite{barkley_cadsmith_2026}. CADSmith \cite{barkley_cadsmith_2026} provides a
multi-agent text-to-CAD solution that generates CadQuery
\cite{cadquery_contributors_cadquery_2026} scripts from natural-language prompts
and refines them through nested correction loops. It is composed of an inner
loop that catches Python and OpenCASCADE
\cite{noauthor_open-cascade-sasocct_2026} execution errors and an outer loop
validates generated geometry against exact measurements. These measurements
(bounding-box dimensions, volume, solid validity) are extracted from the
OpenCASCADE kernel, supplemented by an independent vision-language model judge.
Retrieval-augmented generation over the CadQuery API documentation supplies
type-correct code patterns without fine-tuning. On a 100-prompt benchmark
spanning three difficulty tiers, CADSmith raised execution rate from 95\% to
100\% and reduced mean chamfer distance from 28.37 to 0.74 against a zero-shot
baseline, demonstrating that programmatic geometric validation resolves a class
of dimensional errors that visual-only feedback cannot.


\section{Background}

\subsection{High Power Rocketry}
High power rocketry (Class 2) is an intermediate category of rocketry with a
total impulse ceiling of 40,960 N $\cdot$ s
\cite{federal_aviation_administration_requirements_2008}. Access to respective
impulse ranges (\ref{sec:motor_impulse_by_class}) are granted through
certifications offered through organizations such as Tripoli Rocketry
Association (TRA) and National Association of Rocketry (NAR), both work with the
Federal Aviation Administration (FAA) who administers flight waivers. This class
of rockets is most accessible to adult hobbyists where motors are permitted for
purchase with the appropriate certifications. Advanced high power rocketry
(Class 3) covers rockets beyond 40,960 N $\cdot$ s and up to 889,600 N $\cdot$ s
and requires additional permission from the respective governing bodies
\cite{federal_aviation_administration_requirements_2008}.

\subsubsection{Stability Calculation}

Stability calculation is a critical component during the design process. Rockets
with low stability are more likely to tumble during launch, however an excessive
stability value is prone to directional changes from environmental factors such
as crosswind in a phenomenon called weathercocking \cite{stine_handbook_2004}.
For these reasons, a stability value between 1.00 cal and 1.25 cal is desired.
Stability is determined with three main variables: Center of Pressure (CP),
Center of Gravity (CG), and body tube diameter \cite{stine_handbook_2004}.
Locations for the center of gravity $x_{CG}$ and center of pressure $x_{CP}$ are
measured from the nose tip to the tail and distance between the two is divided
by the body tube diameter $d$ to provide the stability value in the unit of
calibers (Eq. \ref{eq:stability_cg_cp}) \cite{stine_handbook_2004}.

\begin{equation}
    \label{eq:stability_cg_cp}
    Stability = \frac{x_{CG} - x_{CP}}{d}, \quad x_{CG} = \frac{\sum_{i} m_i x_i}{\sum_{i} m_i}, \quad x_{CP} = \frac{\sum_{i} C_{N_i} x_i}{\sum_{i} C_{N_i}}
\end{equation}

The center of gravity is weighted average of the individual mass components
$m_i$ and their respective distance from the nose tip $x_i$
\cite{stine_handbook_2004}. The center of pressure utilizes the Barrowman
\cite{barrowman_practical_2013} equation which considers the individual normal
force coefficients $C_{N_i}$ of each respective component to calculate the
weighted average \cite{stine_handbook_2004}. During the development of the
rocket, the calculated stability often fluctuates on the basis of the recorded
mass of each manufactured component. In addition to the structural components of
the rocket, consideration needs to be given to the motor's change in mass during
launch as this often shifts the center of gravity towards the nose tip.

\subsubsection{Component Design, Manufacturing, and Assembly}

The design and manufacturing of the various rocket components are key phases
foundational to successful launch and recovery events
\cite{stine_handbook_2004}. These phases take into consideration the various
events that occur during and after the launch event such as the placement of
rail buttons along the airframe and the internal pressure caused by recovery
deployment charges. Thoughtful consideration during this design and
manufacturing phase is critical to the successful recovery of the rocket after
launch.

Weight is often the primary constraint within the design process; however, other
considerations such as accessibility, manufacturing, and material are also
non-trivial factors \cite{stine_handbook_2004, bhat_aerospace_2018}. Material
selection is often settled early in the design phase as it establishes initial
constraints of weight, manufacturing, and size \cite{bhat_aerospace_2018}. These
pertain to parts such as body tubes, fins, nose cone, and other internal
components with common material candidates of cardboard, wood, fiberglass,
carbon fiber, and metal alloys \cite{stine_handbook_2004, bhat_aerospace_2018}.
Decisions regarding material selection are often tied with considerations
towards manufacturing as each material presents its own set of manufacturing
challenges.

The assembly of manufactured components is often the final phase before launch
preparation. At this phase, the manufacturing tolerance of components is
assessed as each part is required to fit with their respective mates to ensure
correct flight behavior. Parts that are out of specified tolerances need to be
adjusted through either subtractive or additive modifications to fit correctly
within the assembly. In addition, flight parameters such as the actual center of
gravity are recorded from the rocket assembly, providing real values to adjust
flight simulations.

\subsection{Agentic Systems}

An agentic system is a platform where an LLM is able to autonomously complete
multi-step goals by means of context, reasoning, tool calls, and optimization
\cite{pak_agentic_2026, jadhav_llm-3d_2025, ock_adsorb-agent_2026}. Agentic
systems are useful for complex tasks as the system is capable of operating
beyond the constraints of a single input prompt and the domains of the training
data \cite{nakano_webgpt_2021, yao_react_2023}. These systems are primarily
composed of three main components: LLM enabled reasoning and orchestration, tool
calls for precise and accurate responses, and an agent harnesses which provides
the runtime for the LLM and tools. Together this combination of components
provide the ability to ingest natural language task descriptions, decompose and
execute subtasks, and validate the results.

\subsubsection{Large Language Models}

Large Language Models (LLMs) are transformer-based neural networks trained on
trillions of text tokens through next-token prediction
\cite{vaswani_attention_2017}. Transformers use self-attention to calculate
weighted relationships between tokens of a sequence in parallel. Next-token
prediction is the process of generating a probability distribution of vocabulary
given a sequence of tokens. The model parameters are updated to maximize the
likelihood of the ``true" next token \cite{vaswani_attention_2017}. Tokens are
sub-word units that are made from byte-pair encoding \cite{gage_new_1994}. After
pretraining, these models undergo post-training consisting of supervised
fine-tuning on instruction-response pairs and subsequent reinforcement learning
from human feedback or direct preference optimization to create
instruction-following generations \cite{liu_visual_2023, shi_continual_2025}. The
generation process itself is autoregressive, which means that at inference time
the model samples one token at a time from the output distribution and appends
it to the input. This process repeats until the end of a sequence or a
predefined maximum length.

The context limit is the maximum number of tokens that an LLM can perform
attention upon \cite{vaswani_attention_2017}. This parameter is a key
architectural factor during the development of the model and fixed during
training. Modern frontier models have context windows ranging from 128k beyond
1M tokens. Compute scales quadratically with sequence length
$O(n^2)$ in standard attention, resulting in increasing cost and latency for
longer prompts \cite{vaswani_attention_2017}. Liu et al.\cite{liu_lost_2024}
showed that accuracy on retrieval-style tasks degrades to near or below
performance on shorter contexts when the relevant information is in the middle
of the window. Within agent harnesses handling large amounts of contextual
information, subagents and tool calling can be used to ensure consistent
responses and behavior. 

\subsubsection{Tool Calling}
Tool calling allows for LLMs to interact with their external environment with
predetermined functions. \cite{yao_react_2023, schick_toolformer_2023,
nakano_webgpt_2021}. Early examples of this include WebGPT
\cite{nakano_webgpt_2021} which utilized a simple set of tools to navigate and
collect information through various websites through the guidance of an LLM.
Within the process, the LLM is given a context schema that provides details on
how each tool should be called and in what context it should be called upon.
Since these tools are often snippets of pregenerated code, this allow LLMs to
produce deterministic outputs by invoking external code. The LLM itself is not
capable of executing the code, rather it relies on the higher order function
performing LLM inferences to execute functions based on parsing the generated
response \cite{yao_react_2023}. The tool call then generates a response that is
then returned to the LLM in the form of additional context to the conversation
\cite{pak_agentic_2026, jadhav_llm-3d_2025, ock_adsorb-agent_2026}. Here the LLM
acts as an orchestrator for tools calls, utilizing its reasoning and context to
make informed decisions \cite{yao_react_2023}. Within tool calling there are
failure modes such as schema mismatches and semantic errors which can be avoided
through use of a standard protocol such as the Model Context Protocol (MCP).

\subsubsection{Agent Harnesses}

An agent harness provides the runtime for LLM inference and other
functionalities such as tool calls and user interaction \cite{yao_react_2023,
yang_swe-agent_2024}. The harness is responsible for managing conversation
history, executing tool calls, providing system access, context window
management (summarizing or removing previous turns), and termination conditions
\cite{noauthor_anthropicsclaude-code_2026,
noauthor_google-geminigemini-cli_2026, noauthor_openaicodex_2026}. A common
workflow involves the agent receiving user instructions, passing these to the
LLM, executing necessary tool calls, appending outputs to the conversation, and
repeating until a termination condition is met. Common terminal based agent
harnesses include Claude Code, OpenCode, Antigravity CLI, and Codex CLI which
provide a terminal based user interface for interacting directly with the LLM.

Within agent harnesses, further functionalities include skills, subagents, and
hooks. These features allow for a level of customization enabling further
control of the agentic system. A subagent is a child agent instance that
maintains its own context window, system prompt, and tool subset, when
instantiated by the parent agent \cite{wu_autogen_2024}. Subagents are often
invoked for sub-tasks that work within their own context window and prevents the
parent agent's context window from pollution during subagent task exploration
\cite{wu_autogen_2024}. A summary of a subagent's final response is then
appended to the conversation and used in subsequent inferences.



\section{Methodology}

\subsection{Software Tools}

The core of RocketSmith utilizes three main software tools: OpenRocket
\cite{niskanen_openrocket_2015},
\texttt{build123d}\cite{roger_maitland_build123d_2025}, and PrusaSlicer
\cite{noauthor_prusa3dprusaslicer_2026}. OpenRocket provides the design and
simulation tools to generate rocket blueprints, \texttt{build123d} offers the
capability to generate parameteric \texttt{STEP} files of the declared design
dimensions, and PrusaSlicer calculates more concrete weight estimations and tool
path files for the generated components. With the data obtained during this
pipeline, adjustments are made within the OpenRocket platform and the updates
are made using the downstream software tools. The agentic system primarily
relies on these three main software tools to enable the intelligent automation
of component design and manufacturing.

\subsubsection{OpenRocket}

OpenRocket \cite{niskanen_openrocket_2015} is a design and simulation tool
commonly used for the development of high powered rockets. This platform
provides a database of commonly used rocket components such as solid motors,
body tubes, parachutes, nose cones, and other launch accessories useful for
designing the build of a rocket \cite{niskanen_openrocket_2015}. These
components along with their respective properties such as mass, dimensions, and
material are utilized during the design procedure and provide valuable insight
into flight metrics such as stability and apogee. The design is utilized for the
basis of flight simulations that predict the altitude, velocity, acceleration
and other flight characteristics using variables such as wind conditions,
recovery deployment, and launch rail length. These flight simulations are
valuable as they provide a general estimation of the course of the flight test
before any of the components are built.

As components are manufactured, various dimensional and weight estimations often
deviate slightly from those prescribed in OpenRocket
\cite{niskanen_openrocket_2015}. These initial values can be overridden by
actual values to utilize accurate variables during the generation of flight
simulations. Other factors such as multiple stages or recovery device
deployments can be adjusted on the software platform for further control in more
complex configurations. During development, OpenRocket
\cite{niskanen_openrocket_2015} is often used as the design blueprint for any
CAD, manufacturing, or assembly task.

\begin{figure}
    \centering
    \includegraphics[width=\linewidth]{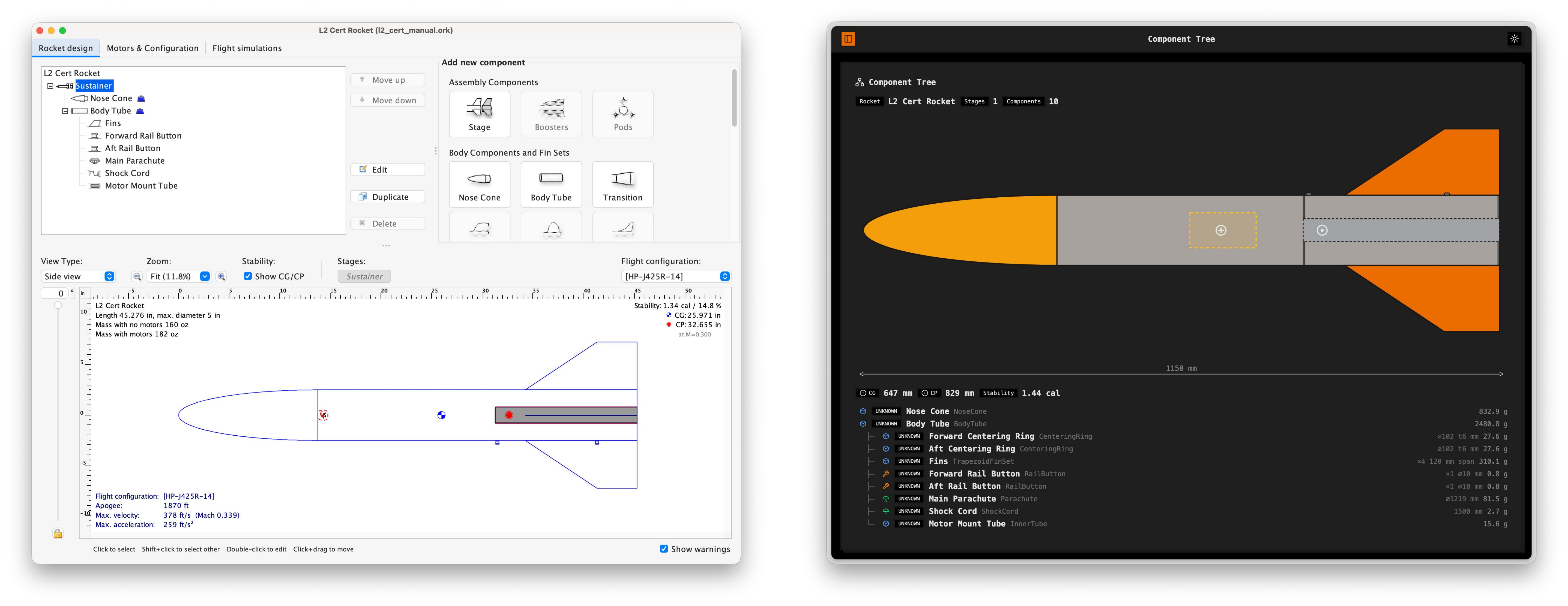}
    \caption{
        (Left) OpenRocket \cite{niskanen_openrocket_2015} provides the
        foundational toolkit for generating rocket designs and running flight
        simulations. (Right) RocketSmith establishes a component tree to use for
        the generation of \texttt{STEP} file components and design related
        skills and subagents.
    }
    \label{fig:openrocket}
\end{figure}

RocketSmith uses OpenRocket \cite{niskanen_openrocket_2015} as the primary
source of truth during the course of rocket development (Figure
\ref{fig:openrocket}). All dimensional, weight, and assembly configurations are
first implemented within OpenRocket \cite{niskanen_openrocket_2015} before being
sent to any downstream software tools. Integration of the OpenRocket
\cite{niskanen_openrocket_2015} software into the agentic system is enabled
using the OpenRocket Helper \cite{noauthor_openrocketorhelper_2026} package
which provides Python bindings to the Java based OpenRocket
\cite{niskanen_openrocket_2015} platform. This allows the agentic system to
utilize the database of existing components, create and modify components,
perform overrides for mass and dimensions, and run flight simulations. The
designs and simulations built using OpenRocket \cite{niskanen_openrocket_2015}
makes this software foundational to the reliable generation of flight capable
rocket designs by the RocketSmith agentic system.

\subsubsection{\texttt{build123d}}

\texttt{build123d} \cite{roger_maitland_build123d_2025} is a parametric CAD
Python library that wraps the OpenCASCADE
\cite{noauthor_open-cascade-sasocct_2026} geometric kernel and uses
context-manager and operator-overload conventions to define solids imperatively.
This package is a successor to CADquery
\cite{cadquery_contributors_cadquery_2026} and shares the same kernel, but
exposes a more Pythonic authoring surface that maps cleanly onto the linear,
top-to-bottom structure an LLM tends to produce, with named intermediate state
at each step. Code-based formulation is adopted over generative-3D approaches
(diffusion, point-cloud or mesh transformers) for four reasons specific to
agentic authoring. Firstmost, the script is deterministic and parametric:
identical inputs yield identical geometry, and the LLM can target a specific
feature for adjustment (changing a single fillet radius or shifting one fin
chord) without regenerating the entire shape, in contrast to diffusion
approaches that must resample the whole solid for any local edit. Second, the
output is a \texttt{STEP} file whose exact bounding box, volume, and mass (via
material density) can be extracted programmatically and checked against the
manifest's expected values, making closed-loop verification possible without
lossy visual proxies. Third, \texttt{STEP} is the industry exchange format, and
flows directly into the slicer (PrusaSlicer) and any downstream finite-element
or CAM tooling without the mesh-cleanup steps that generative outputs typically
require. Fourth, \texttt{build123d} failures surface as Python exceptions
carrying a geometric reason, so the agent can read the traceback and adapt
rather than producing silently invalid geometry.

\begin{figure}
    \centering
    \includegraphics[width=\linewidth]{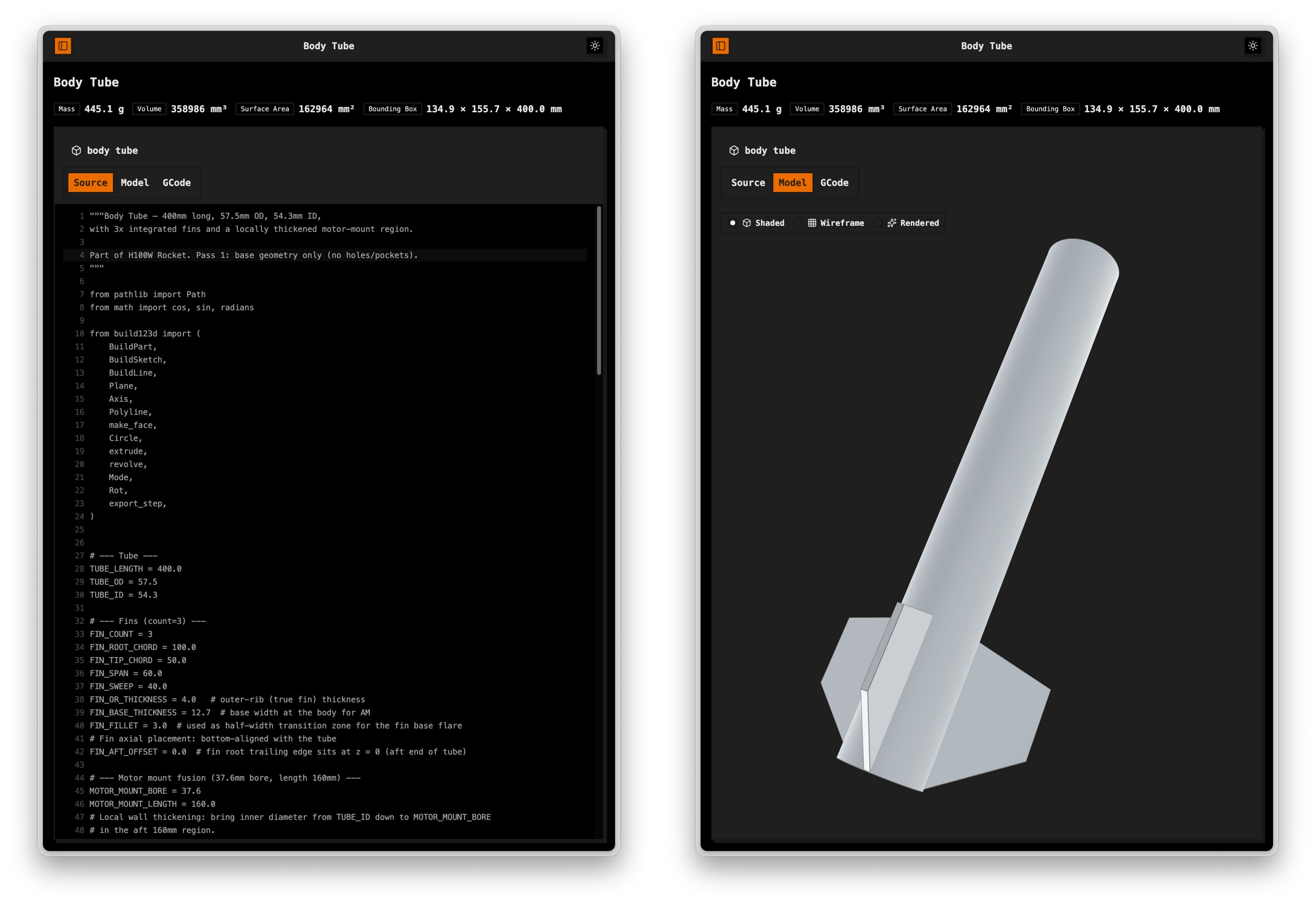}
    \caption{
        (Left) \texttt{build123d} is the primary framework to use for generating
        the parametric part configurations written in Python. (Right)
        \texttt{STEP} file is generated from running Python snippet and
        visualization allows for easy part adjustment.
    }
    \label{fig:cadsmith}
\end{figure}

Within RocketSmith, \texttt{build123d} is invoked once stability and flight
readiness has been verified in the OpenRocket design and simulation phase. The
CAD agent receives the finalized part list with each component's dimensions and
authors a \texttt{build123d} script for every printable part. As the scripts
execute and \texttt{STEP} files are written, the corresponding parts appear
progressively in the project GUI (Figure \ref{fig:cadsmith}), allowing the user
to observe build progress and issue iterative adjustments to the agent (for
example, requesting a thicker wall, a longer shoulder, or a re-positioned vent).
When all parts are complete, the GUI presents an assembly view rendering every
\texttt{build123d} part in its correct relative orientation. The user can
iterate back and forth with the agent on fine adjustments or alterations until
satisfied, after which the final \texttt{STEP} files are exported for slicing
and additive manufacturing.

\subsubsection{PrusaSlicer}
\begin{figure}
    \centering
    \includegraphics[width=\linewidth]{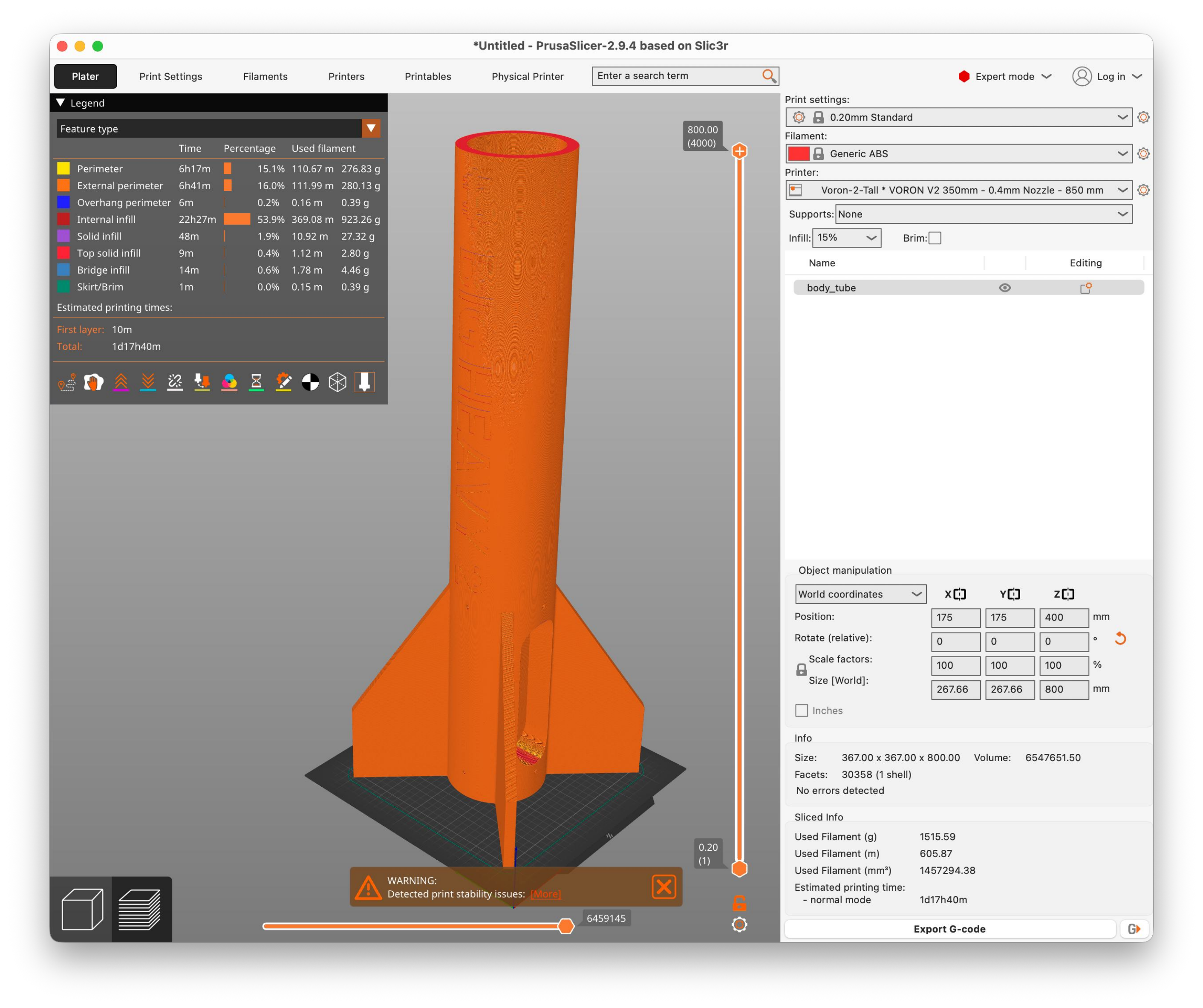}
    \caption{
        Weight estimation for generated \texttt{STEP} files for various
        components are obtained using Prusaslicer
        \cite{noauthor_prusa3dprusaslicer_2026} using configurations for the
        expected material.
    }
    \label{fig:prusaslicer}
\end{figure}

PrusaSlicer \cite{noauthor_prusa3dprusaslicer_2026} is a configurable CAD model
slicer software forked from the Slic3r \cite{noauthor_slic3rslic3r_2026} project
for Fused Deposition Modeling (FDM) tool paths. The slicer provides configurable
parameters for three main sections: print settings, filaments, and printers.
Configurations within print settings determine build specific settings regarding
layer height, tool speed, infill, supports, and other print related settings.
Material specific settings can be assigned in the filament section allowing for
adjustment of bed temperature, hotend temperature, and various cooling
parameters. The printers section provides machine specific specifications that
need consideration during the slicing process such as dimensional limitations
and number of available extruders. Within RocketSmith, PrusaSlicer (Figure
\ref{fig:prusaslicer}) allows for a more precise weight estimation of the
generated \texttt{STEP} file component through slicing with the expected
material, print, and printer configurations. Once sliced, the generated tool
paths are uploaded to the respective printer for manufacturing where once
fabricated are weighed to obtain actual weight overrides.

\subsection{Agentic System}

The development of high powered rockets is augmented with the usage of an
agentic system which is capable of operating in zero-shot
\cite{wei_finetuned_2021, kojima_large_2022} and human-in-the-loop
\cite{mosqueira-rey_human---loop_2023} conditions. Under both operating
conditions the system provides a graphical user interface to monitor and
evaluate the outputs including CAD models, flight simulations, and component
trees. The use of subagents allows for the compartmentalization of specialized
context specific to operations concerning the OpenRocket, PrusaSlicer, CADSmith
\cite{barkley_cadsmith_2026}, and other RocketSmith functionality. Skills
provide a concrete workflow for tool usage regarding specific behaviors that are
common throughout the duration of the system.

\subsubsection{Graphical User Interface (GUI)}

\begin{figure}
    \centering
    \includegraphics[width=\linewidth]{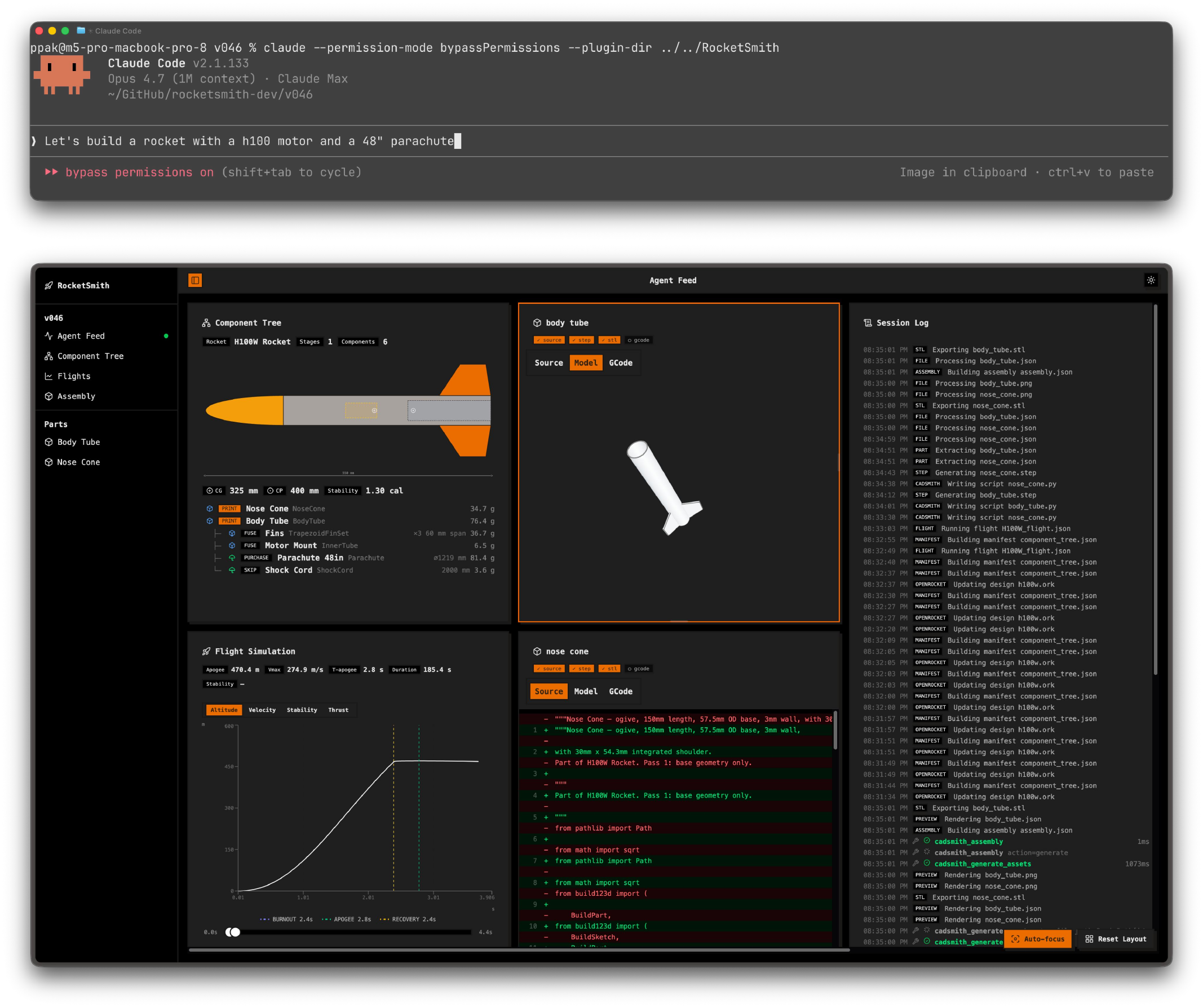}
    \caption{
        (Top Window) Terminal with Claude Code provides the primary means to
        control the RocketSmith agentic system with initial and follow up
        prompts for designing high powered rockets. (Bottom Window) Web based
        GUI displays tool calls results, flight simulations, component trees,
        and CADSmith model outputs. Locally deployed server to connects two
        windows with realtime updates allowing for visualization of key aspects
        of the rocket development process.
    }
    \label{fig:gui}
\end{figure}

The Graphical User Interface (GUI) provides a medium to observe the actions of
the agentic system during the rocket development process, especially useful in
design related tasks regarding assembly and modeling. The main dashboard of the
GUI enables a holistic view into the various functions of the RocketSmith
agentic system, highlighting the current active card the agent is concerned with
(Figure \ref{fig:gui}). These cards include specific aspects of the agentic
system including the rocket component tree, flight simulation results, CADSmith
\cite{barkley_cadsmith_2026} scripts and models, and assembly configurations. At
its core, the GUI simply a ``readonly" visualization of the current state of the
agentic system as the primary method of executing actions is through the CLI
based agent harness of Claude Code \cite{noauthor_anthropicsclaude-code_2026}.

\subsubsection{Subagents}

The agentic system is composed of 6 individual subagents: \texttt{rocketsmith},
\texttt{cadsmith}, \texttt{gui}, \texttt{manufacturing}, \texttt{openrocket},
and \texttt{prusaslicer}. Each individual subagent is responsible for a limited
scope of the agentic system and enables efficient utilization of context. The
\texttt{rocketsmith} subagent is responsible for the overall function of the
agentic system denoting the project file structure layout, declaring subagent
scopes, and applying various soft guardrails. Subagents for \texttt{openrocket},
\texttt{cadsmith}, and \texttt{prusaslicer} provide additional information
regarding how to best utilize each respective software platform's API and when
it is appropriate to do so. The \texttt{manufacturing} subagent manages the hand
off steps between the OpenRocket \cite{niskanen_openrocket_2015} and CADSmith
\cite{barkley_cadsmith_2026} software programs evaluating potential Design for
Additive Manufacturing (DFAM) and Design for Manufacturing (DFM) adjustments.
The \texttt{gui} subagent manages the visualizations shown to the user,
navigating between different pages and highlighting active cards.

\subsubsection{Skills}

Skills provides a concrete set of instructions to provide the agent when
performing a specific task and in RocketSmith a total of 7 different skills are
instantiated: \texttt{design-for-additive-manufacturing},
\texttt{generate-structures}, \texttt{mass-calibration},
\texttt{modify-structures}, \texttt{motor-selection},
\texttt{print-preparation}, and \texttt{stability-analysis}. Within workflows of
the RocketSmith agentic system, \texttt{motor-selection} is the first skill that
is executed as this provides a base to build the designs and structures upon.
Afterwards OpenRocket \cite{niskanen_openrocket_2015} is performed to generate
a stable design which is then verified with the \texttt{stability-analysis}
skill. An optional \texttt{design-for-additive-manufacturing} skill concerned
with component consolidation is performed at this stage, specifically
investigating motor mounts and couplers whose parts can be combined into a
single 3D printable assembly. The skill for \texttt{generate-structures} is
performed to outline the CAD model for the first pass (i.e. fins, walls, etc.)
and a subsequent skill of \texttt{modify-structures} is executed to add minor
adjustments (i.e. tap holes, screw holes, etc.). The last
\texttt{print-preparation} outlines the procedure to send the \texttt{STEP}
files for each component over to PrusaSlicer
\cite{noauthor_prusa3dprusaslicer_2026} for weight estimation and tool path
generation.

\subsection{Manufacturing}

\begin{figure}
    \centering
    \includegraphics[width=\linewidth]{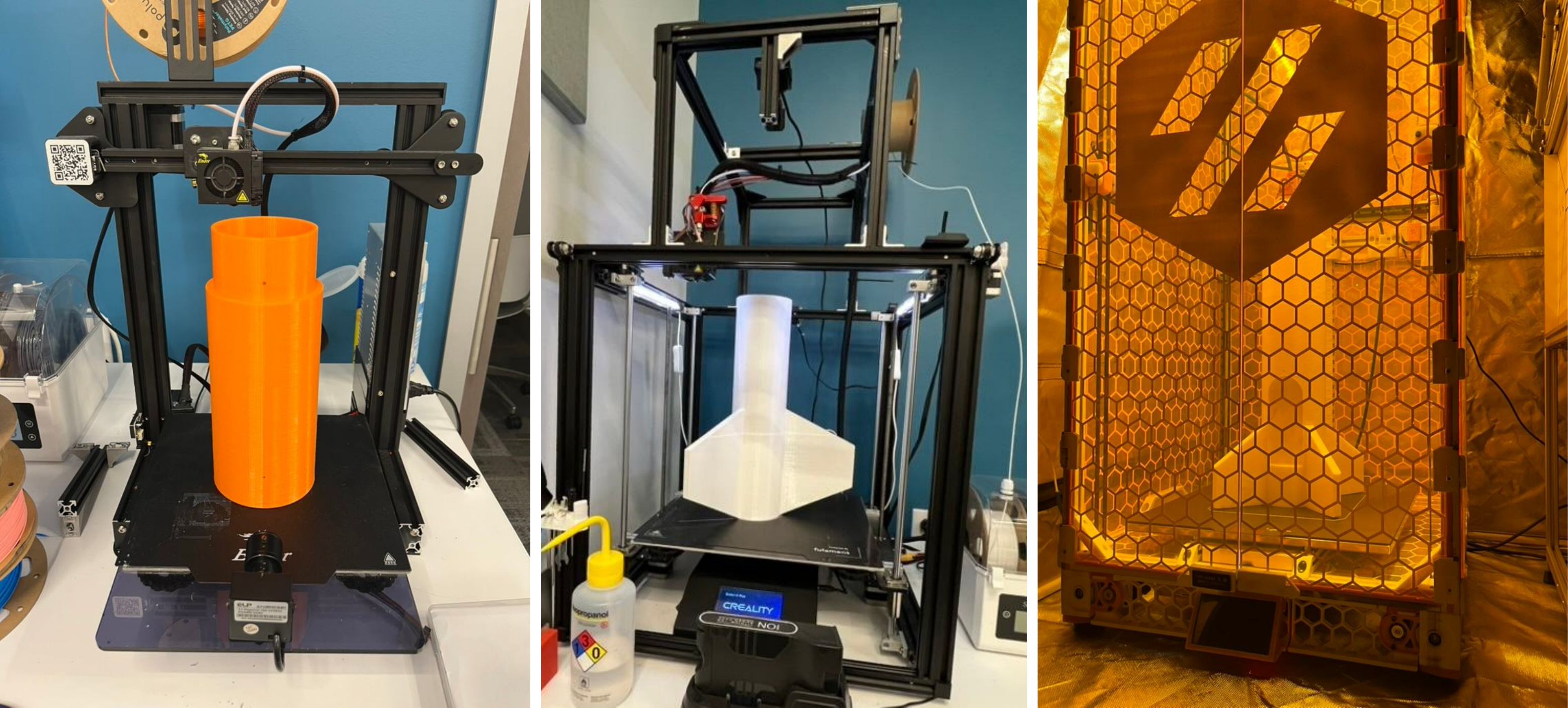}
    \caption{
        (Left) Ender 3 with printed middle airframe component for High Power 2.
        (Middle) Ender 5 Plus with printed lower airframe component for High
        Power 2. (Right) \texttt{Voron-2-Tall}
        \cite{pak_ppak10voron-2-tall_2026} with printed lower airframe component
        for High Power 1.
    }
    \label{fig:printers}
\end{figure}

A total of four high powered rockets were manufactured from designs generated
using the RocketSmith agentic system; referred to as High Power 1, 2, 3, and 4
(Figure \ref{fig:rocket_diagrams}). All airframe components were fabricated with
the Fused Deposition Modeling (FDM) additive manufacturing process in either
Polyethylene Terephthalate Glycol (PETG) \cite{szykiedans_selected_2017} or
Acrylonitrile Butadiene Styrene (ABS) \cite{tanikella_tensile_2017} filament. A
variety of FDM printers (Figure \ref{fig:printers}) were utilized to print the
various components including a Creality Ender 3, Creality Ender 5 Plus, and the
custom built \texttt{Voron-2-Tall} \cite{pak_ppak10voron-2-tall_2026}
(\ref{sec:voron-2-tall}). The materials of PETG and ABS were selected as opposed
to Polylactic Acid (PLA) \cite{tymrak_mechanical_2014} for its relatively higher
heat deflection temperatures.

All components were printed with a 15\% gyroid infill, 4 to 5 vertical walls,
and a 0.20 mm layer height. Three of the four high power rockets were composed
of two manufactured airframe components, those being the nose cone and airframe
body. These airframe bodies extended past the printable z height of the Ender 5
Plus (400 mm) and were instead printed using the \texttt{Voron-2-Tall}
\cite{pak_ppak10voron-2-tall_2026} with ABS filament. The dimensions of these
nose cones were sufficiently small enough to manufacture using the Ender 5 Plus
and printed with PETG filament. High Power 2 is an outlier as it was designed to
be fabricated with commercial 3D printers with components dimensioned to fit
within the build volumes of an Ender 5 Plus and Ender 3 printers. A combination
of these two printers were used to print the nose cone, middle airframe, and
lower airframe using PETG filament. During the manufacturing process
consideration was given to the mating of the various components and necessary
adjustments were made to ensure adequate fit between the various components.

\subsection{Assembly}

\begin{figure}
    \centering
    \includegraphics[width=\linewidth]{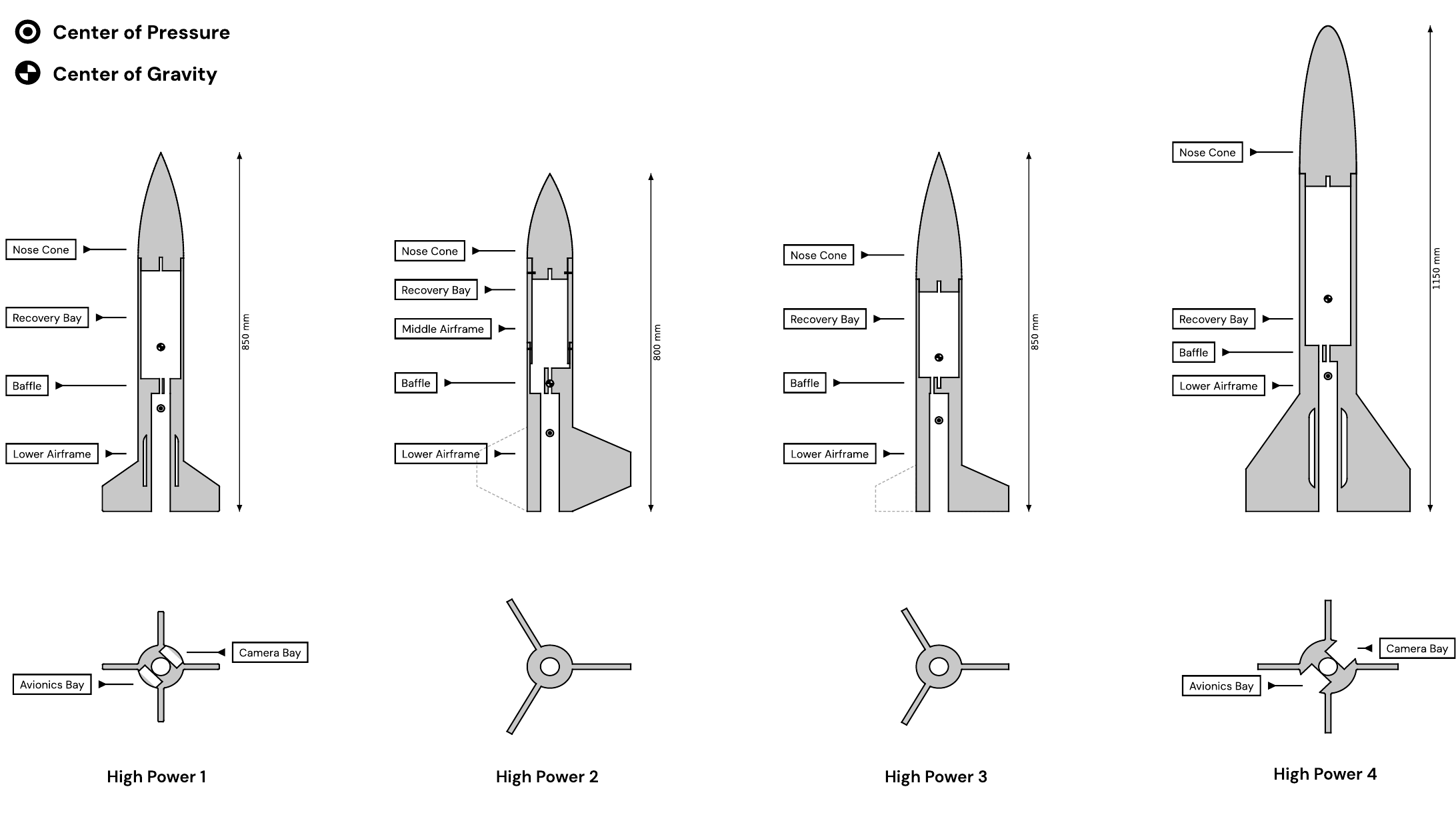}
    \caption{
        Rocket design schematics for High Power 1 (Pak), 2 (Loghmani), 3
        (Barkley), and 4 (Pak) along with their respective components and bays.
        The recovery bay for High Power 2 is constructed using a separate middle
        airframe component in order to fit onto Creality Ender 5 Plus print bed.
        High Power 1 and 4 include slots to house altimeters and cameras for
        data recording during flight testing.
    }
    \label{fig:rocket_diagrams}
\end{figure}

Figure \ref{fig:rocket_diagrams} displays four high powered rockets that were
designed and assembled by their respective engineers: Pak, Loghmani, and
Barkley. Pak designed and constructed two separate high power rockets: High
Power 1 (Section \ref{sec:assembly_hp1}) with a level 1 impulse classification
and High Power 4 (Section \ref{sec:assembly_hp4}) with a level 2 impulse
classification. Loghmani designed and built High Power 2 (Section
\ref{sec:assembly_hp2}), a level 1 impulse high powered rocket designed for
manufacturing using the commercially available Creality Ender 5 Plus and
Creality Ender 3 printers. Barkley also designed and assembled a level 1 impulse
high powered rocket, High Power 3 (Section \ref{sec:assembly_hp3}), with
simplified designs enabled through large format FDM printing. Further
characteristics and quantities for each rocket are outlined in Table
\ref{tab:rocket_quantities}.

\begin{table}[ht]
    \centering
    \caption{Specific characteristics and quantities for High Power 1, 2, 3, and 4}
    \label{tab:rocket_quantities}
    \begin{tabular}{lcccc}
        \hline
        \textbf{Quantity} & \textbf{High Power 1} & \textbf{High Power 2} & \textbf{High Power 3} & \textbf{High Power 4} \\
        \hline
        Developer         & Pak             & Loghmani      & Barkley       & Pak \\
        Motor             & H100W           & H219T         & H219T         & J425R \\
        Recovery          & 36" Parachute   & 48" Parachute & 48" Parachute & 48" Parachute \\
        Altimeter         & StratoLogger CF & N/A           & N/A           & StratoLogger CF \\
        Camera            & RunCam 5        & N/A           & N/A           & RunCam 5 \\
        Length            & 850~mm          & 800~mm        & 850~mm        & 1150~mm   \\
        Diameter          & 101.6~mm        & 101.6~mm      & 101.6~mm      & 127.0~mm  \\
        Fins              & 4               & 3             & 3             & 4 \\
        Printed Parts     & 4               & 3             & 2             & 6 \\
        Materials         & ABS \& PETG     & PETG          & ABS \& PETG   & ABS \& PETG \\k        Shear Pins        & 0               & 4             & 0             & 0 \\
        Mass              & 2249.8~g        & 2166.09~g     & 1857~g        & 5158.8~g  \\
        Mass (dry)        & 1988.8~g        & 1905.09~g     & 1596~g        & 4527.8~g  \\
        Stability         & 1.26~cal        & 1.63~cal      & 1.27~cal      & 1.52~cal  \\
        Stability (dry)   & 1.46~cal        & 1.82~cal      & 1.47~cal      & 1.68~cal  \\
        \hline
    \end{tabular}
\end{table}

\subsubsection{High Power 1}
\label{sec:assembly_hp1}

\begin{figure}
    \centering
    \includegraphics[width=\linewidth]{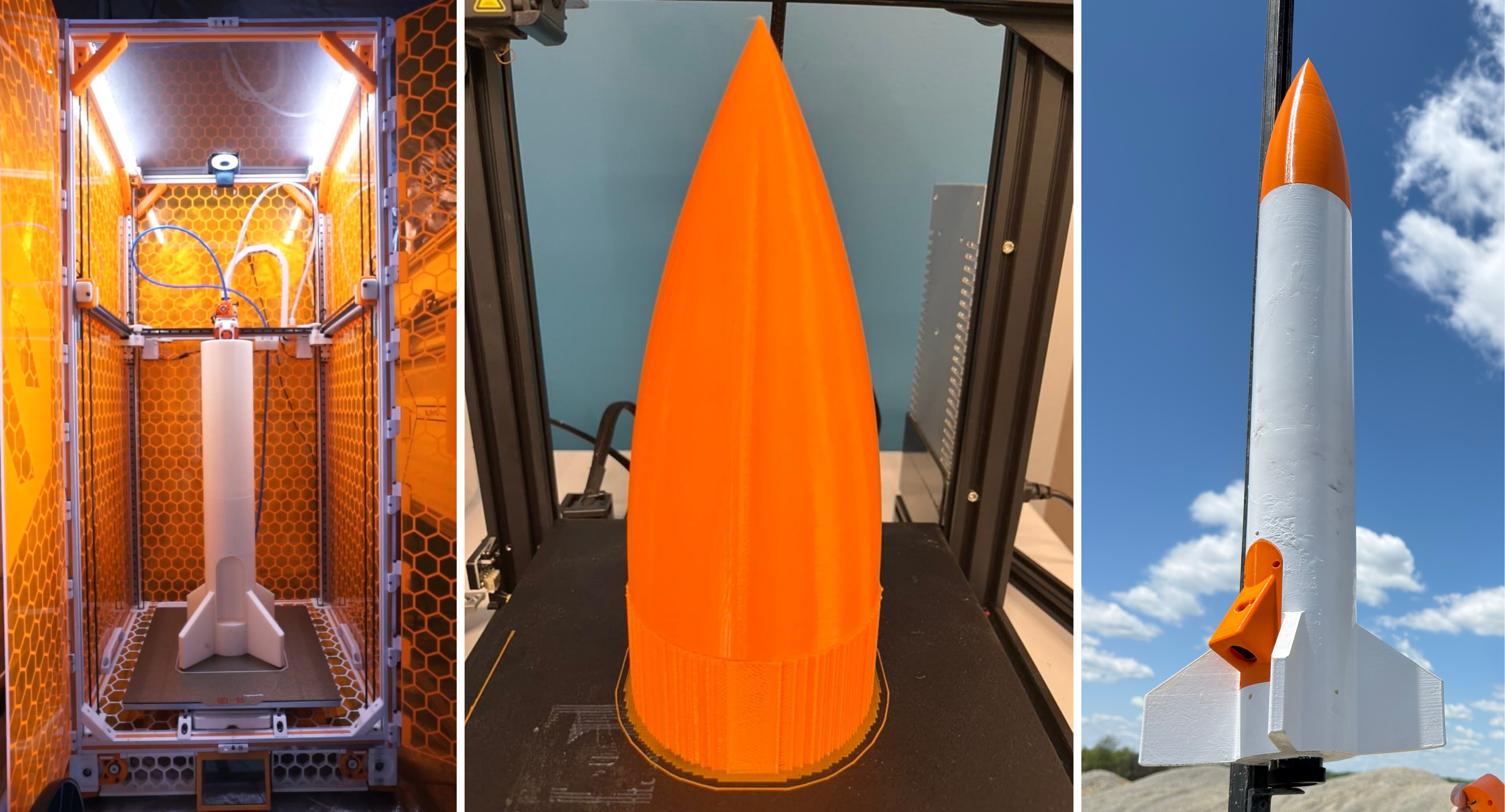}
    \caption{
        (Left) Finished lower airframe printed using ABS filament over the
        course of several days (Middle) Nose cone printed using PETG with
        supports to account for the overhang above the shoulder (Right)
        Assembled \textit{v38} rocket prepared for launch on 1010 rail.
    }
    \label{fig:assembly_hp1}
\end{figure}

High Power 1 (Figure \ref{fig:assembly_hp1}) designed and constructed by Pak, is
a single motor deploy level 1 high powered rocket composed of two primary
components of a lower airframe and nose cone. This rocket, \textit{v38} short
for version 38 produced with RocketSmith, was designed around the use of a 38 mm
AeroTech H100W solid rocket motor and recovery with a 36" parachute. A pair of
5/16" steel eyebolts were attached to the base of the nose cone and the center
of the baffle in the lower airframe, providing anchor points for the kevlar
shock cord used to connect the recovery parachute to the rest of the body. The
lower airframe was printed with \texttt{Voron-2-Tall} FDM printer
\cite{pak_ppak10voron-2-tall_2026} using ABS filament with slots cut out for
StratoLogger CF altimeter and a RunCam 5 camera for flight data recording. The
nose cone was printed with the Creality Ender 5 Plus using PETG filament.

\subsubsection{High Power 2}
\label{sec:assembly_hp2}

\begin{figure}
    \centering
    \includegraphics[width=\linewidth]{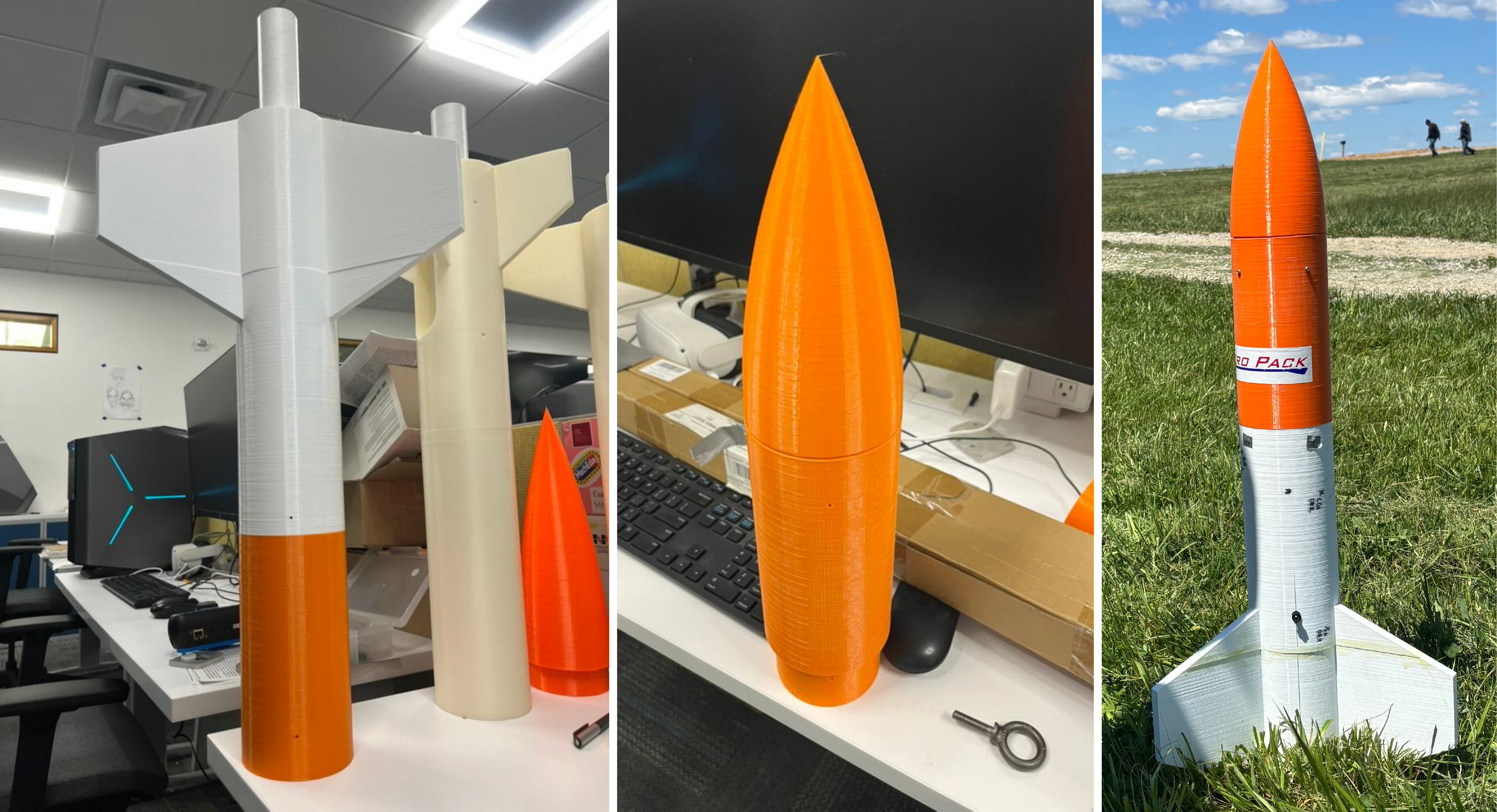}
    \caption{
        (Left) 38 mm motor tube inserted into the lower airframe to ensure
        adequate fit (Middle) Assembled nose cone and middle airframe components
        secured with M3 bolts and heat inserts (Right) Completely assembled
        rocket at launch site.
    }
    \label{fig:assembly_hp2}
\end{figure}

High Power 2, \textit{Ruminator}, (Figure \ref{fig:assembly_hp2}) is a
single-stage high power rocket with a three component assembly designed and
constructed by Loghmani. It used an AeroTech H219T solid rocket motor and a 48"
parachute. During development, RocketSmith was given the motor specification and
a three-body airframe requirement, and from this generated a trapezoidal
three-fin design and inner diameter constraints. The three parts consist of the
lower airframe, middle airframe, and nose cone. The rocket was 31.5 inches in
length and the body's outer diameter was 4 inches with a 0.425 inch wall
thickness. The ogive nose cone and middle airframe were each 7.87 inches in
length, and the lower airframe was 15.75 inches in length. The nose cone and
middle airframe were designed to be printed on a Creality Ender 3, and the lower
airframe on a Creality Ender 5 Plus. The lower airframe had a three-legged
baffle design to hold the eyebolt that was attached to the parachute, and the
nose cone similarly had a hole in the bottom to hold an eyebolt. A shock cord
was attached to each eyebolt with a 48" parachute attached and packed inside the
middle airframe for deployment at apogee. The three-part design simplified
eyebolt installation. The baffle's gap allowed the apogee ignition charge to
separate the nose cone and middle airframe parts from the lower airframe. To
ensure the nose cone and middle airframe stayed attached, heat inserts and
screws were added. The lower airframe and middle airframe were connected via
four shear pins, designed to shear when the apogee ignition charge activates.
RocketSmith was used to adjust the CAD model and create pilot holes for the heat
inserts, shear pins, and eyebolts. To join all parts, RocketSmith also added
1.97 inch shoulders between the lower and middle airframes, and between the
middle airframe and the nose cone.

\subsubsection{High Power 3}
\label{sec:assembly_hp3}

\begin{figure}
    \centering
    \includegraphics[width=\linewidth]{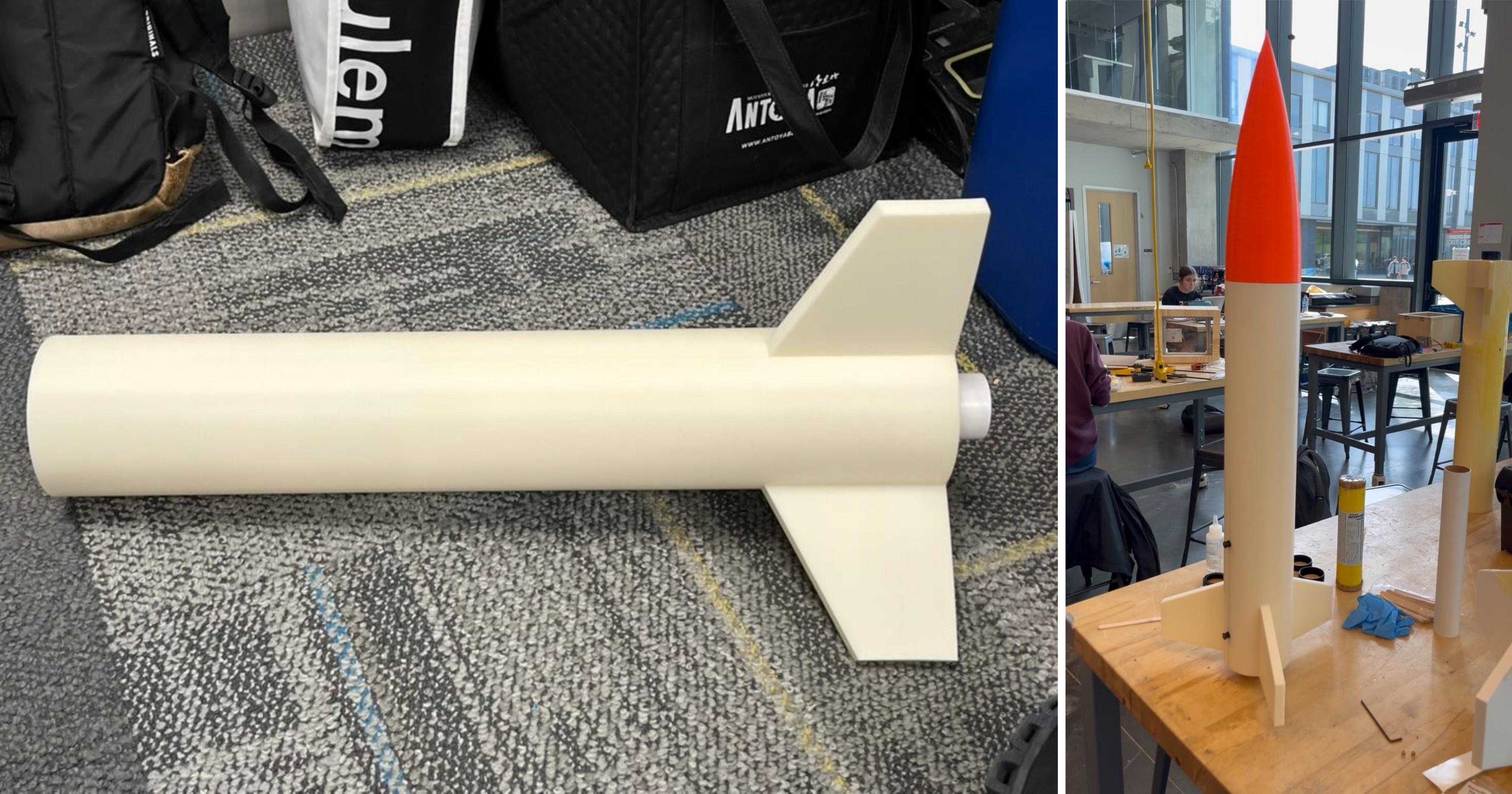}
    \caption{
        (Left) Lower airframe printed using ABS filament with motor tube
        inserted (Right) Initial completed assembly of rocket with post design
        installation of rail buttons.
    }
    \label{fig:assembly_hp3}
\end{figure}

High Power 3 (Figure \ref{fig:assembly_hp3}), the \textit{H219T Thunderbolt},
was the simplest of the case studies, comprised of only a nose cone and a single
main airframe rather than the segmented nose-cone, middle-airframe, and
lower-airframe arrangement used in High Power 2. The body tube was around 4
inches (101.6 mm) in outer diameter with a quarter-inch (6.35 mm) wall printed
in ABS, chosen for its higher heat resistance during motor ignition relative to
PLA or PETG. An AeroTech H219T solid rocket motor seated in a printed
motor-mount region whose wall was locally thickened and integrated into the main
airframe rather than authored as a separate inner tube. The nose cone was
retained by a friction-fit shoulder rather than by shear pins as this method
applies less resistance during motor ejection. For recovery anchoring, the
CADSmith \cite{barkley_cadsmith_2026} agent placed a pilot hole at the center of
the underside of the nose cone shoulder and a second pilot hole at the center of
the integrated ejection-charge baffle inside the main airframe. A 5/16" steel
eyebolt was attached to both ends of the rocket providing a connection to the
lower airframe and nose cone via shock cord. A 48 inch parachute was attached
and packed forward of the baffle for recovery deployment.

\subsubsection{High Power 4}
\label{sec:assembly_hp4}

\begin{figure}
    \centering
    \includegraphics[width=\linewidth]{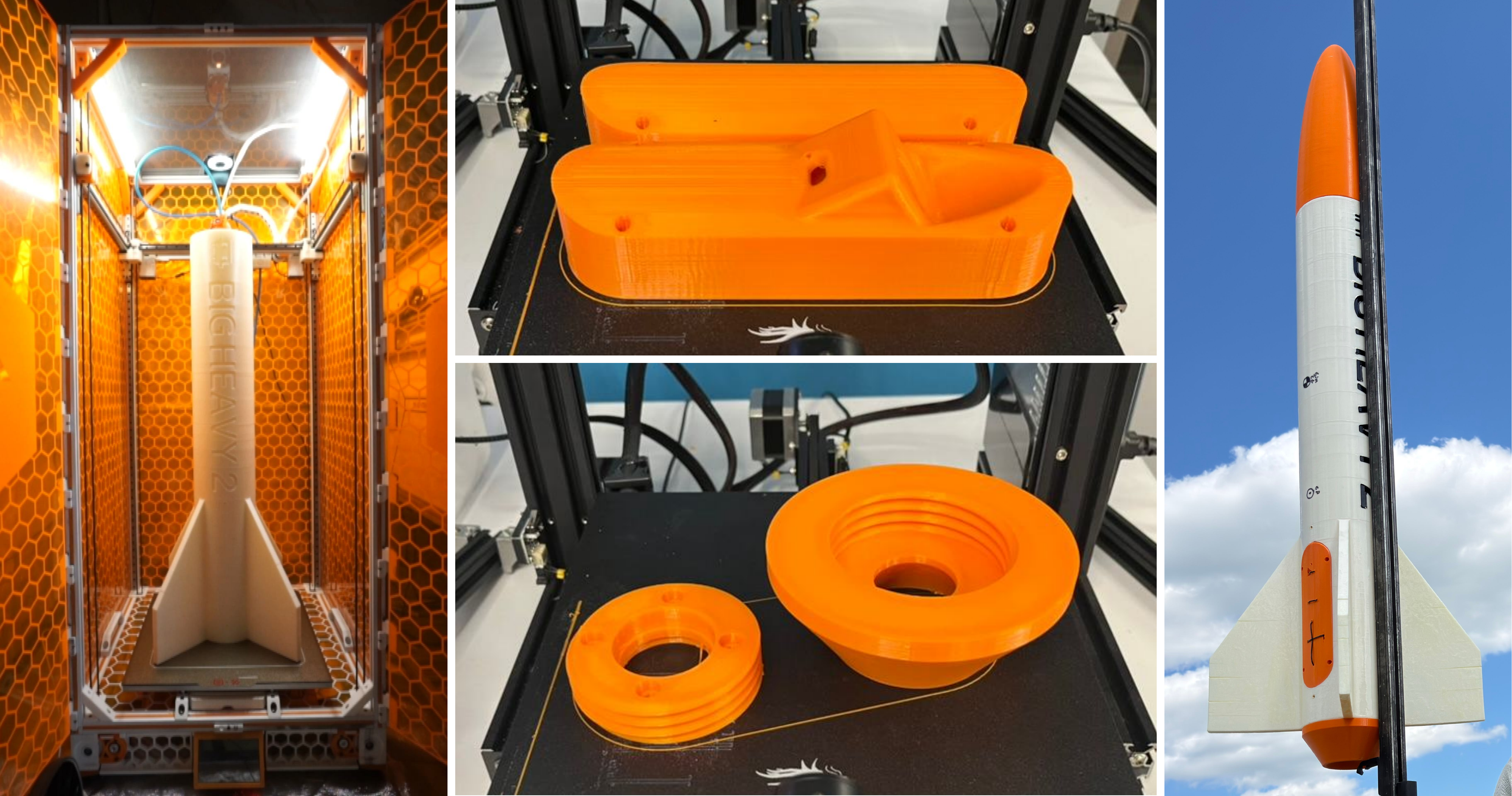}
    \caption{
        (Left) Finished lower airframe component printed with
        \texttt{Voron-2-Tall} \cite{pak_ppak10voron-2-tall_2026} using ABS
        filament over a duration of 4 days (Top Middle) Covers for altimeter and
        camera components along with (Bottom Middle) motor retainer components
        printed with PETG filament using Creality Ender 3 printer (Right)
        Assembled \textit{Big Heavy 2} prepared on launch rail.
    }
    \label{fig:assembly_hp4}
\end{figure}

High Power 4 (Figure \ref{fig:assembly_hp4}), \textit{Big Heavy 2}, is a level 2
impulse class rocket designed around an AeroTech J425R single use solid rocket
motor. Of the four rockets, it is the largest with a lower airframe height of
800 mm and an overall height of 1150 mm. Similar to High Power 1, it has slots
cut out to house a StratoLogger CF altimeter and a RunCam 5 camera within the
lower airframe. The rocket utilizes a 48" parachute for the recovery phase and
utilizes two steel eyebolts to connect the nose cone and lower airframe via
kevlar shock cord. The covers for the electronics along with the motor retainers
are designed and printed outside the scope of the RocketSmith agentic system and
printed with PETG on the Creality Ender 3 printer.

\section{Results}

\subsection{Flight Tests}
Flight tests of all High Power rockets were performed on Sunday, May 3rd 2026 at
\textit{Dragon's Fire Field}, an hour drive south of Pittsburgh in Fayette
County. Clear sky and moderate winds (around 10 mph) from the west enabled
nearly ideal flight conditions during the permitted operating window of 9 AM to
5 PM (\ref{sec:launch_site_weather_conditions}). All flight tests were performed
under the supervision of members from the Pittsburgh prefecture of the Tripoli
Rocketry Association. Preflight checks of all rockets were done by the Range
Safety Officer (RSO) and all suggested adjustments were applied accordingly
before flight testing.

\begin{table}[h]
    \centering
    \caption{
        Expected maximum altitude, velocity, and acceleration values along with
        measured data points for rockets with altimeters (High Power 1 and High
        Power 4).
    }
    \label{tab:flight_data}
    \begin{tabular}{llcccc}
        & & \multicolumn{4}{c}{\textbf{High Power}} \\
        \cline{3-6}
        & & \textbf{1} & \textbf{2} & \textbf{3} & \textbf{4} \\
        \hline
        \multirow{2}{*}{\textbf{Apogee}}     & Expected & 338.8~m   & 334~m     & 473.9~m   & 570.1~m   \\
                                      & Measured & 276~m     & N/A       & N/A       & 479~m     \\
        \hline
        \multirow{2}{*}{\textbf{Max Velocity}}     & Expected & 76.4~m/s  & 94.1~m/s  & 114.5~m/s & 115.2~m/s \\
                                      & Measured & 63.3~m/s  & N/A       & N/A       & 92.4~m/s  \\
        \hline
        \multirow{2}{*}{\textbf{Max Acceleration}} & Expected & 48.4~m/s\textsuperscript{2} & 128.4~m/s\textsuperscript{2} & 147.7~m/s\textsuperscript{2} & 78.8~m/s\textsuperscript{2} \\
                                      & Measured & 26.9~m/s\textsuperscript{2} & N/A       & N/A       & 54.8~m/s\textsuperscript{2} \\
        \hline
    \end{tabular}
\end{table}

High Power 1 - 3 were simultaneously launched within the same volley with High
Power 4 launched by itself on a later volley (Figure \ref{fig:results_launch}).
Of the four rockets, two were successfully recovered in reflyable condition
(High Power 3 and High Power 4) whereas High Power 1 encountered structural
issues from the ejection charge and High Power 2 encountered recovery deployment
issues. Collected data from High Power 1 and High Power 4 showed that both
reached apogees close to that of the predicted flight simulation (80\% and 84\%
of expected altitude respectively).

\begin{figure}
    \centering
    \includegraphics[width=\linewidth]{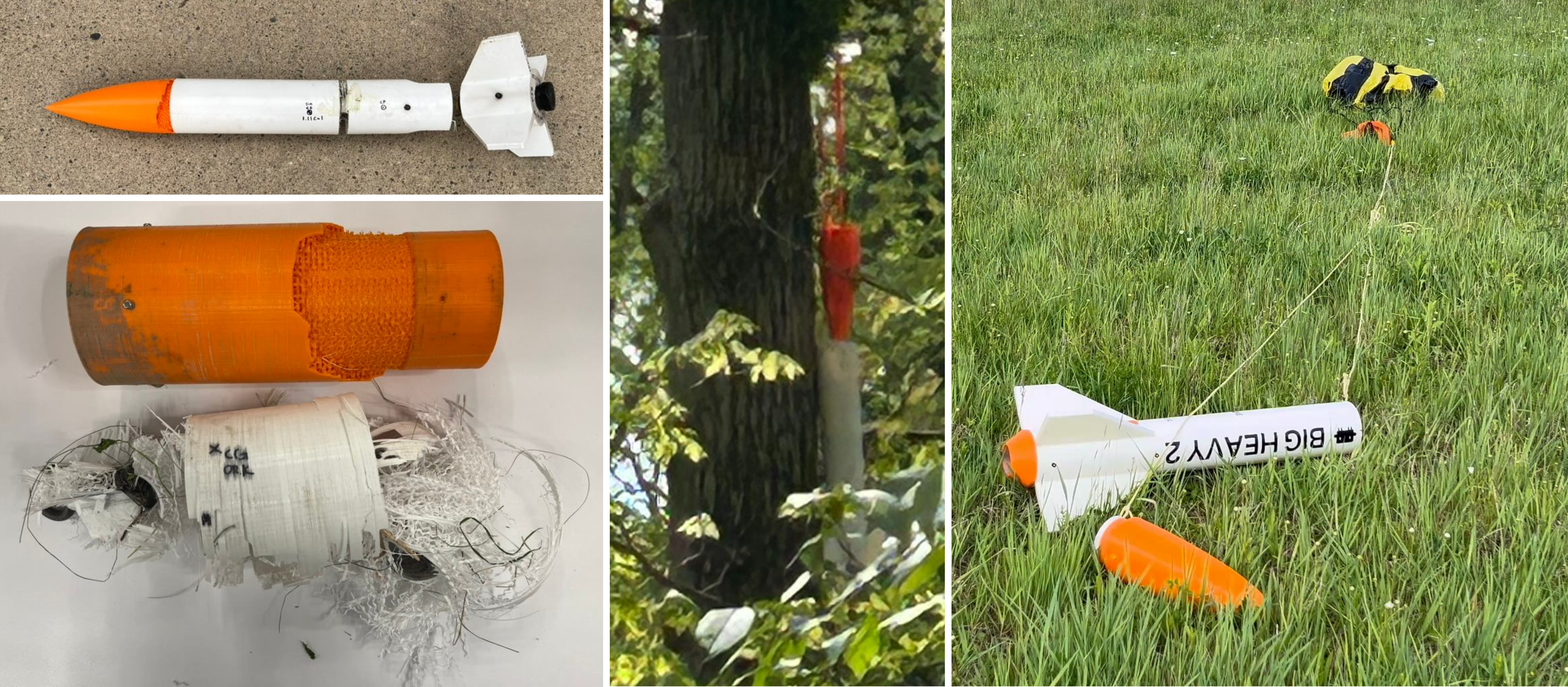}
    \caption{
        (Top Left) Recovered High Power 1 components after fracture from
        ejection charge (Bottom Left) Recovered High Power 2 components after
        recovery bay separation failure (Middle) Successfully recovered High
        Power 3 after landing in tree (Right) High Power 4 after successful
        recovery deployment and landing in field.
    }
    \label{fig:results_recovery}
\end{figure}

\subsubsection{High Power 1}
\label{sec:results_hp1}
\begin{figure}
    \centering
    \includegraphics[width=\linewidth]{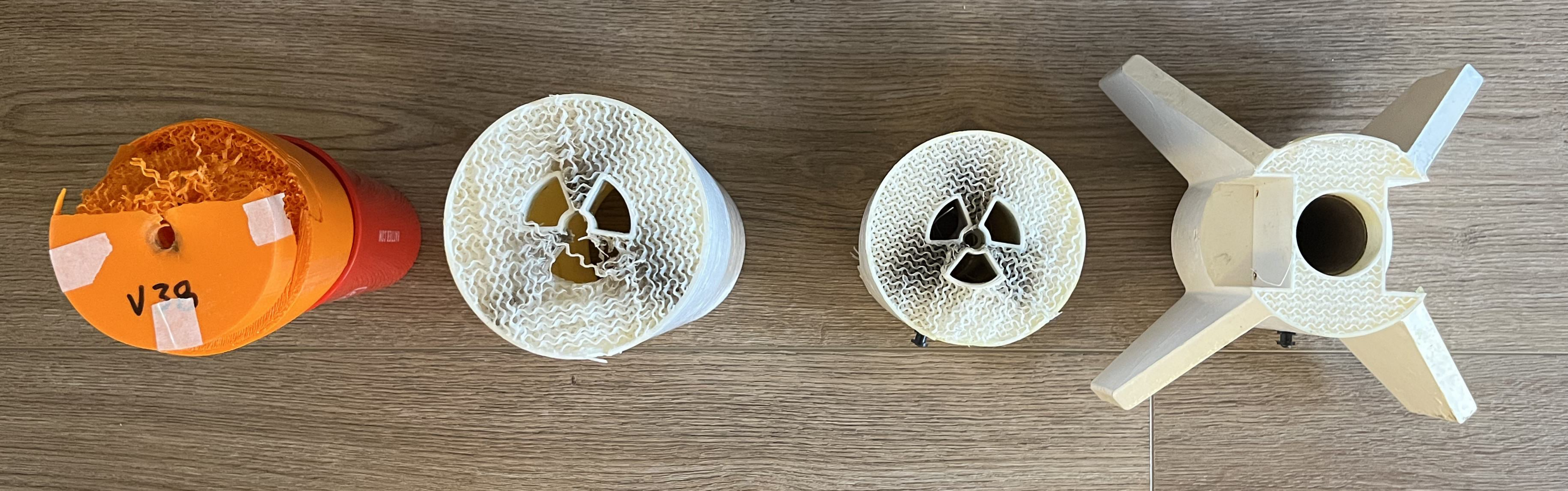}
    \caption{
        Recovered components of High Power 1 after successful launch including:
        (Left to Right) damaged nose cone with eyebolt removed post recovery,
        top section of lower airframe component with split at baffle section,
        middle section of lower airfame split between baffle and fin can, and
        bottom section of lower airframe including housing for electronics.
    }
    \label{fig:recovery_v38}
\end{figure}

High Power 1 was launched successfully in the first volley alongside High Power
2 and High Power 3. Its various components were recovered damaged and in
non-reflyable condition due to a catastrophic structural failure within the
lower airframe. The onboard electronics including the StratoLogger CF altimeter
(Figure \ref{fig:flight_data}) and the RunCam 5 camera (Figure
\ref{fig:runcam_v38}) were successful in recording flight data and recovered in
reusable condition. Data collected from the altimeter shows that the rocket
reached a measured maximum altitude of 276 m (905 ft), around 80\% of its
expected apogee of 338 m (1108 ft). Recovered components of High Power 1 can be
seen in Figure \ref{fig:results_recovery} where a clear split is visible in
sections of the lower airframe midway through the baffle and just above the
motor tube (Figure \ref{fig:recovery_v38}). At recovery it was visible that the
eyebolt of the lower airframe was dislodged from its installed location within
the baffle causing the bottom half lower airframe to return to the ground in
free fall conditions. The top portion of the lower airframe and nose cone landed
gently with the successful recovery deployment of the 36" parachute.

\subsubsection{High Power 2}
\label{sec:results_hp2}
High Power 2 was tested alongside High Power 1 and High Power 3 without onboard
instrumentation. The launch was vertical and visibly stable, without any
weathercocking or tipping during the boost phase. The recovery ejection charge
is believed to have detonated at apogee but separation of the recovery bay did
not occur and the rocket returned to the ground at high velocity. High Power 2
hit the ground with ballistic impact and embedded itself approximately 4 meters
into the soil, embedding components of the nose cone and middle airframe. The
force of landing detached the lower airframe from the middle airframe and nose
cone, and the baffle detached from the lower airframe, lodging itself deep into
the middle airframe and covering the parachute and shock cord. Post-flight
inspection confirmed that the lower airframe's fins remained intact, although
cracks along the body of the lower airframe exposed the internal print infill.


\subsubsection{High Power 3}
\label{sec:results_hp3}

High Power 3 was the first successful launch and recovery of a high powered
rocket designed by RocketSmith. At launch the rocket displayed visual stability
with no off-axis tipping during the boost phase. Although no flight data was
recorded, visually estimated apogee was consistent with what was expected from
the flight simulation. The rocket entered recovery stage shortly after reaching
apogee and experienced significant descent time with the deployed 48" parachute.
Due to the lengthy descent period, the rocket drifted and was lodged in a tree,
however landed intact and in reflyable condition (Figure
\ref{fig:results_recovery}). In order to recover the rocket, the trees
surrounding and holding the rocket were cut and during this retrieval the lower
airframe of the rocket split. To the best of the authors knowledge, the failures
within the lower airframe occurred during the retrieval phase and did not occur
during the flight of High Power 3.

\begin{figure}
    \centering
    \includegraphics[width=\linewidth]{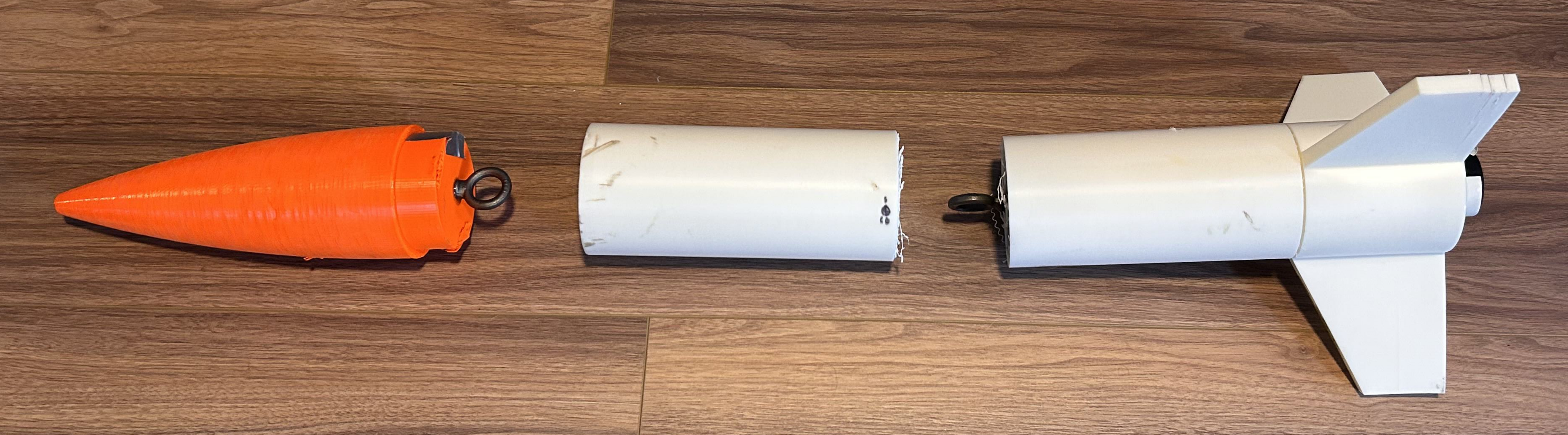}
    \caption{
        Recovered High Power 3 rocket after removal from tree with lower
        airframe split occurring during the retrieval.
    }
    \label{fig:recovery_hp3}
\end{figure}

\subsubsection{High Power 4}
\label{sec:results_hp4}

High Power 4 was the second successful launch and recovery of a level 2 impulse
classification rocket developed using RocketSmith. It was launched in its own
separate volley and recovered successfully in reflyable condition with minimal
effort. The onboard StratoLogger CF altimeter measured an apogee of 479 m (1571
ft), 84\% of the expected 570 m (1870 ft). Separation of the recovery bay
occurred approximately at apogee and deployment of the 48" parachute was
performed successfully. Due to the single deployment of recovery components the
main parachute also carried the components a significant distance where it was
easily recovered in a farm field. Video of the entire flight and descent was
captured using the onboard RunCam 5 camera (Figure
\ref{fig:runcam_big_heavy_2}).

\begin{figure}
    \centering
    \includegraphics[width=\linewidth]{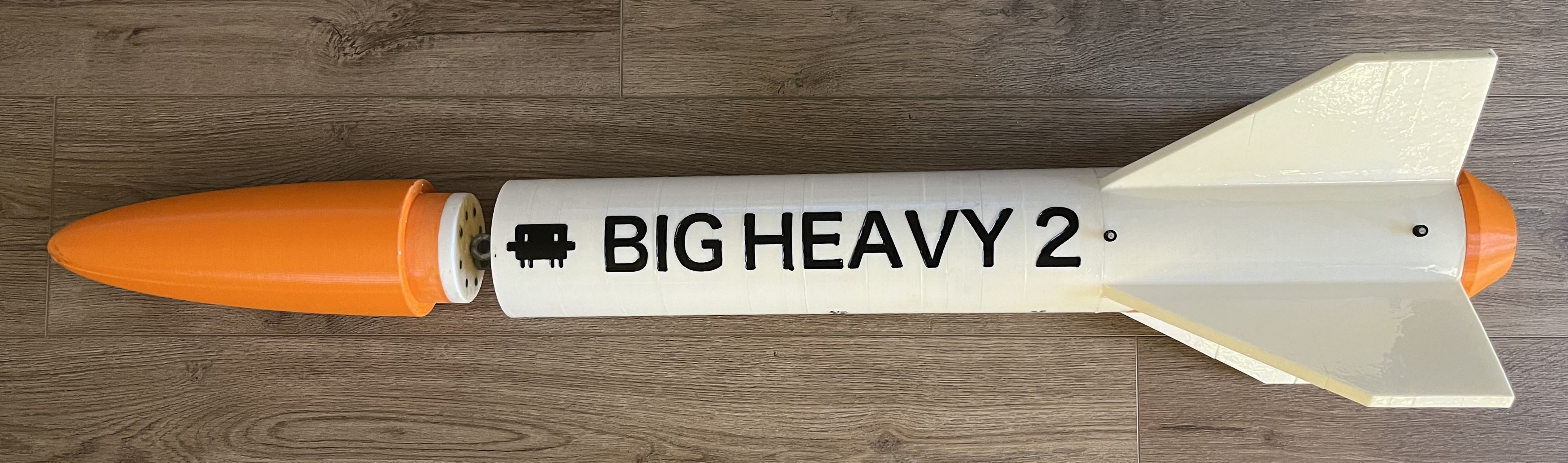}
    \caption{
        Recovered High Power 4 rocket in reflyable condition with minor cosmetic
        blemishes along the nose cone and lower airframe.
    }
    \label{fig:recovery_hp4}
\end{figure}

\section{Discussion}

\subsection{Design and Manufacturing}
RocketSmith enables the automated design and manufacturing of high powered
rockets, however during the development process of the flight tested high power
rockets, various steps required input from the authors. Within the design space,
High Power 1 and High Power 4 utilized manually designed CAD components. The
lower airframe and nose cone components of High Power 1 and High Power 4 were
designed solely with the RocketSmith agentic system with human guidance. Using
RocketSmith, slots for electronics were cut into the lower airframe in
anticipation for covers designed at a later stage. These covers were designed
outside of the RocketSmith agentic system due to relative complexity of the
desired design. In addition, the motor retainer for High Power 4 was designed
separately using SolidWorks and later integrated into the rest of the rocket.

For all rockets, the integrated PrusaSlicer slicer platform was used primarily
for weight estimation of each component. Actual slicing and printing of each
component's \texttt{STEP} file was executed outside the RocketSmith agentic
system with print profiles suitable for the each design and printer (i.e.
inclusion of supports and print speed). This decision to perform manual slicing
reduced the risk of print failures allowing the authors to adhere to the tight
manufacturing schedule before launch. Specific to components printed with ABS,
cracking was observed through various parts of the airframe, especially in High
Power 1 and High Power 4. These structural defects were fixed post build using
epoxy to fill in the gaps and provide more structural integrity. Approaches to
address this issue could utilize in-situ monitoring techniques
\cite{jadhav_llm-3d_2025, pak_thermopore_2024, ogoke_deep_2024,
bostan_accurate_2025} to detect these issues but a more immediate solution would
be to increase the insulation of the build chamber and potentially utilize a
less temperature sensitive material such as PETG. For future versions of
RocketSmith, the utilization of PrusaSlicer will be developed to allow for
greater visual validation and customization, enabling a truly end-to-end
pipeline.

The most time consuming aspect of high power rocket development is the
consistent iteration to designs and simulations to ensure suitable values
regarding stability and recovery deployment. A zero-shot approach is feasible
with RocketSmith, however there are many additional considerations during
development that promote a human in the loop utilization of this agentic system.
Regardless of which approach is taken, this system has shown to significantly
reduce the time and friction of the successful development and testing of high
powered rockets.

\subsection{Flight Tests}

Of flight tests, electronics were installed on two of the four high powered
rockets. These are the rockets that were developed by Pak who had previously
earned a Tripoli Rocket Association Level 1 Certification and best allocated
the limited electronics to maximize the chance of recovery. As such, the
performed flight tests were primarily concerned with evaluating the design and
manufacturing capability of RocketSmith where each rocket was shown to be
successful in achieving stable launch. With instrumentation installed into High
Power 1 and High Power 4, further insight into each of their respective flights
can be extracted.

\begin{figure}
    \centering
    \includegraphics[width=\linewidth]{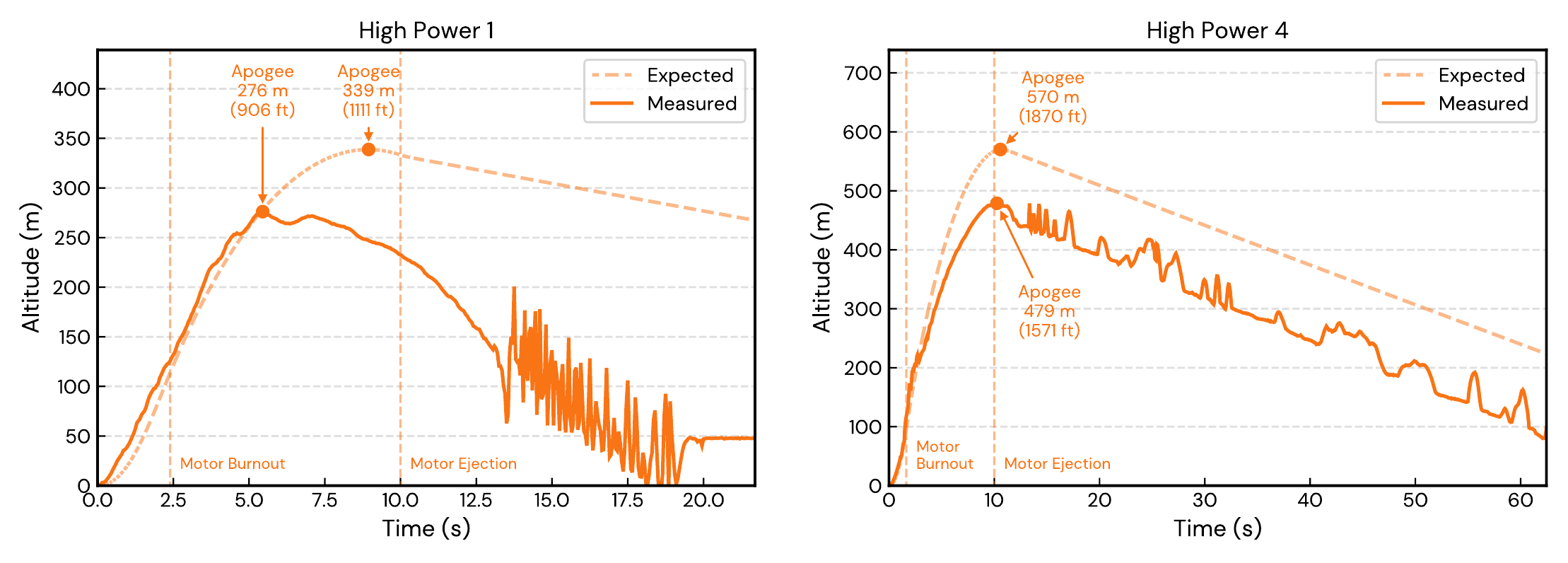}
    \caption{
        Flight data collected from StratoLogger altimeters on High Power 1 and
        High Power 4 plotted alongside expected altitudes. (Left) High Power 1
        reaches around 80\% of expected altitude split of the lower airframe
        resulted in a noisy altitude reading towards the end of flight. (Right)
        High Power 4 reaches approximately 84\% of expected apogee and altimeter
        readings indicate a smooth descent resulting in a successful recovery of
        rocket in reflyable condition.
    }
    \label{fig:flight_data}
\end{figure}

High Power 1 reached around 80\% of its expected apogee, however, the recovery
phase deviates significantly from the expected trajectory (Figure
\ref{fig:flight_data}). From the altimeter and the onboard video recording, the
motor ejection charge detonation was observed around 14 seconds after launch.
This is significantly past the expected motor ejection of 10 seconds just past
the point in time where the rocket is expected to reach apogee. With the delayed
separation of the recovery bay, the rocket experienced significantly higher
forces from the deployment of the 36" parachute attributing to the catastrophic
failure of the lower airframe. Root Cause Corrective Action (RCCA) for this
issue would involve more precise delay charge drill adjustment to reduce
potential deviation at the time of recovery bay separation. Another option would
be to use an altimeter based approach prescribed to deploy recovery devices at a
period just after apogee and further enabling reliability through redundancy.

High Power 4 performed near to the expected flight trajectory with the recovery
bay separation occurring at almost exactly the measured apogee. The measured
apogee is approximately 84\% of the expected with the descent rate matching that
of the expected from flight simulation (Figure \ref{fig:flight_data}). This
enabled an easy successful recovery of the rocket with all components remaining
intact and in reflyable condition. Compared to all the tested high power
rockets, High Power 4 performed the best however did have the most human
involvement during the development.

Performed flight tests show that the manufactured high powered rocket reflect
properties closely to the designs produced by the RocketSmith agentic system.
Flight characteristics such as stability, weight, and apogee are metrics that
are critical to a successful launch, however other considerations regarding
manufacturing and assembly are necessary for a successful recovery. Factors
regarding manufacturing and other specialized high power rocketry knowledge
would further enable the agentic system to construct designs capable of reliable
launch and reuse.

\section{Conclusion}
Flight tests performed using High Power 1, 2, 3, and 4 show that RocketSmith is
capable of the automated development of additively manufactured high powered
rockets. With the use of subagents and skills, the agentic system is able to
outline suitable high power rocket blueprints, design for additive
manufacturing, generate CAD models,  and optimize flight critical values through
subsequent iterations. The outputs from RocketSmith are additively manufactured
with a variety of FDM printers and flight tested at a launch event where two of
the four tested high power rockets were successfully recovered in reflyable
condition. All rockets achieved a stable launch and onboard instrumentation
showed an apogee accuracy of 80\% and 84\% for High Power 1 and High Power 4
respectively. These results show that an agentic system is capable of
successfully designing complex assemblies suitable for additive manufacturing,
validated on domain specific applications such as high powered rocketry.

\clearpage
\appendix

\renewcommand{\thesection}{Appendix \Alph{section}}

\section{Motor Impulse Classification}
\label{sec:motor_impulse_by_class}

Motor classification is determined by the total impulse delivered by the
propulsion system; specifically, this work utilizes a series of solid composite
propellant. Total impulse is defined by the thrust integrated over the burn time
\cite{terry_w_mccreary_phd_experimental_2021}. Equation
\ref{eq:thrust_and_total_impulse} defines the thrust force $F$ in Newtons as the
product of mass flow rate of propellant $\dot{m}$ in $kg/s$ and the exhaust
velocity $v_e$ in $m/s$ \cite{terry_w_mccreary_phd_experimental_2021,
sutton_rocket_2026}. The integral of the thrust force for the burn duration
$t_b$ established the total impulse $I_{total}$ of the motor
\cite{terry_w_mccreary_phd_experimental_2021, sutton_rocket_2026}.

\begin{equation}
\label{eq:thrust_and_total_impulse}
    F = \dot{m} v_e, \quad I_{total} = \int_0^{t_b} F(t) \, dt
\end{equation}

Specific to amateur rocket activities, the Federal Aviation Administration (FAA)
officially designates 3 classes of rockets: Class 1 - Model Rocket, Class 2 -
High Power Rockets, and Class 3 - Advanced High Power Rockets
\cite{federal_aviation_administration_requirements_2008}. Tripoli Rocketry
Association and National Association of Rocketry provide further rocket motor
codes for total impulse ranges (A - O) where the letter specifies the total
impulse, intermediate number specifies the average thrust in newtons and the last
number specifies the time delay between motor burnout and recovery ejection
(i.e. a motor with designation H100W-14 defines a motor with a total within the
range of 160.01 - 320 Newton-seconds, an average thrust of 100 newtons, a
manufacturer specific white color, and a maximum time delay of 14 seconds).
Table \ref{tab:motor_impulse_by_class} in \ref{sec:motor_impulse_by_class}
outlines the various motor classifications by total impulse range which double
with each subsequence letter code and organization specific certification levels
(1 - 3) are required to purchase each respective high powered motor. Most
amateur rocketry activities are designated as Class 2 High Power Rockets which
limits motors to a combined total impulse of 40,960 Newton-seconds (H to O
motors) \cite{federal_aviation_administration_requirements_2008,
stine_handbook_2004}.

Tripoli Rocketry Association and National Association of Rocketry require proof
of certification for the purchasing and launching of high powered rockets.
Certifications are granted for successful launch and recovery of a rocket under
the organization's specified conditions \cite{stine_handbook_2004}.

\begin{table}[ht]
    \centering
    \caption{Rocket motor classification by total impulse (Classes A--O).}
    \begin{tabular}{ccc}
        \hline
        \textbf{Class} & \textbf{Total Impulse (N$\cdot$s)} & \textbf{Certification Required} \\
        \hline
        A & 1.26 -- 2.50   & None \\
        B & 2.51 -- 5.00   & None \\
        C & 5.01 -- 10.0   & None \\
        D & 10.01 -- 20.0  & None \\
        E & 20.01 -- 40.0  & None \\
        F & 40.01 -- 80.0  & None \\
        G & 80.01 -- 160   & None \\
        H & 160.01 -- 320  & Level 1 \\
        I & 320.01 -- 640  & Level 1 \\
        J & 640.01 -- 1{,}280  & Level 2 \\
        K & 1{,}280.01 -- 2{,}560 & Level 2 \\
        L & 2{,}560.01 -- 5{,}120 & Level 2 \\
        M & 5{,}120.01 -- 10{,}240 & Level 3 \\
        N & 10{,}240.01 -- 20{,}480 & Level 3 \\
        O & 20{,}480.01 -- 40{,}960 & Level 3 \\
        \hline
    \end{tabular}
    \label{tab:motor_impulse_by_class}
\end{table}
\clearpage

\section{\texttt{Voron-2-Tall}}
\label{sec:voron-2-tall}

\texttt{Voron-2-Tall} \cite{pak_ppak10voron-2-tall_2026} is a modified Voron 2.4
\cite{noauthor_vorondesignvoron-2_2026} FDM printer with an extended z axis
build height maximum of 930 mm, particularly useful for printing tall parts such
as rocket airframe components. This machine was built from a kit with original
parts source for a Voron 2.4 \cite{noauthor_vorondesignvoron-2_2026} with
dimensions of 350 mm x 350 mm x 330 mm. Custom 2020 aluminum extrusions (1130
mm), MGN9H linear rails (1000 mm), GT2 belts, and enclosure panels (483 mm x
1103 mm x 3 mm) were sourced to extend the z axis over a range of 1000 mm. With
the application of frame braces, the build height limit is reduced from 930 mm
to 850 mm. Additional modifications were made including the incorporation of a
\textit{DragonBurner} toolhead for a smaller form factor along with the usage of
CANBUS for the reduction of wires and removal of cable chains.

\begin{figure}
    \centering
    \begin{minipage}{0.32\linewidth}
        \includegraphics[width=\linewidth]{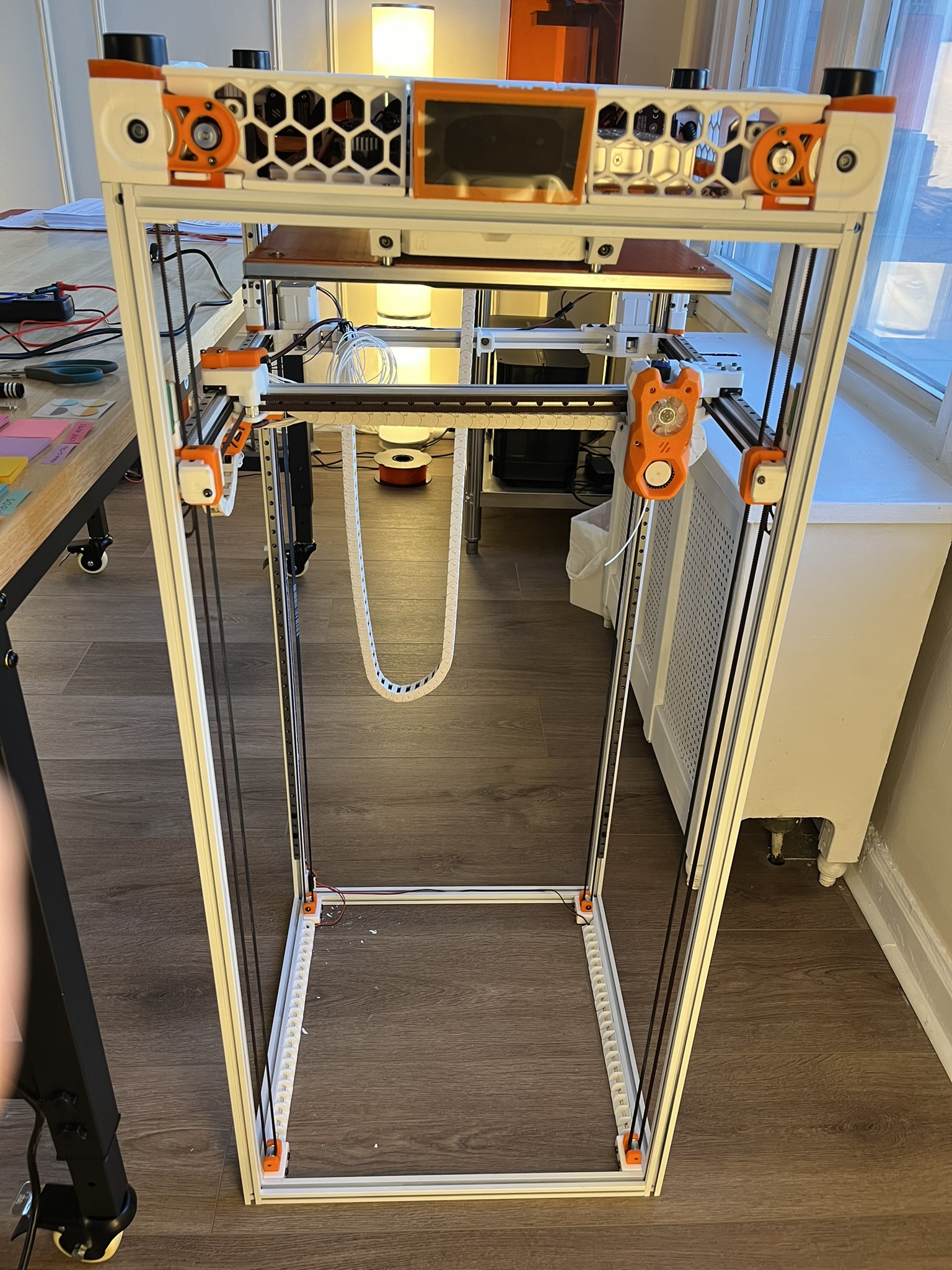}
    \end{minipage}
    \hfill
    \begin{minipage}{0.32\linewidth}
        \includegraphics[width=\linewidth]{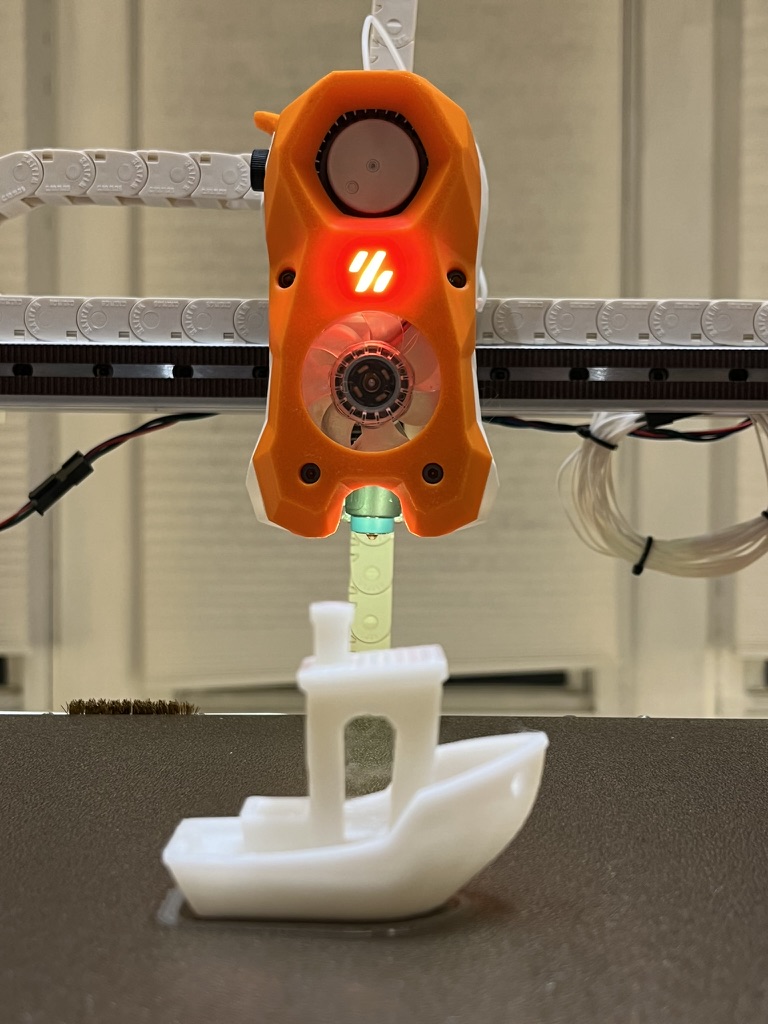}
    \end{minipage}
    \hfill
    \begin{minipage}{0.32\linewidth}
        \includegraphics[width=1.33\linewidth, angle=-90]{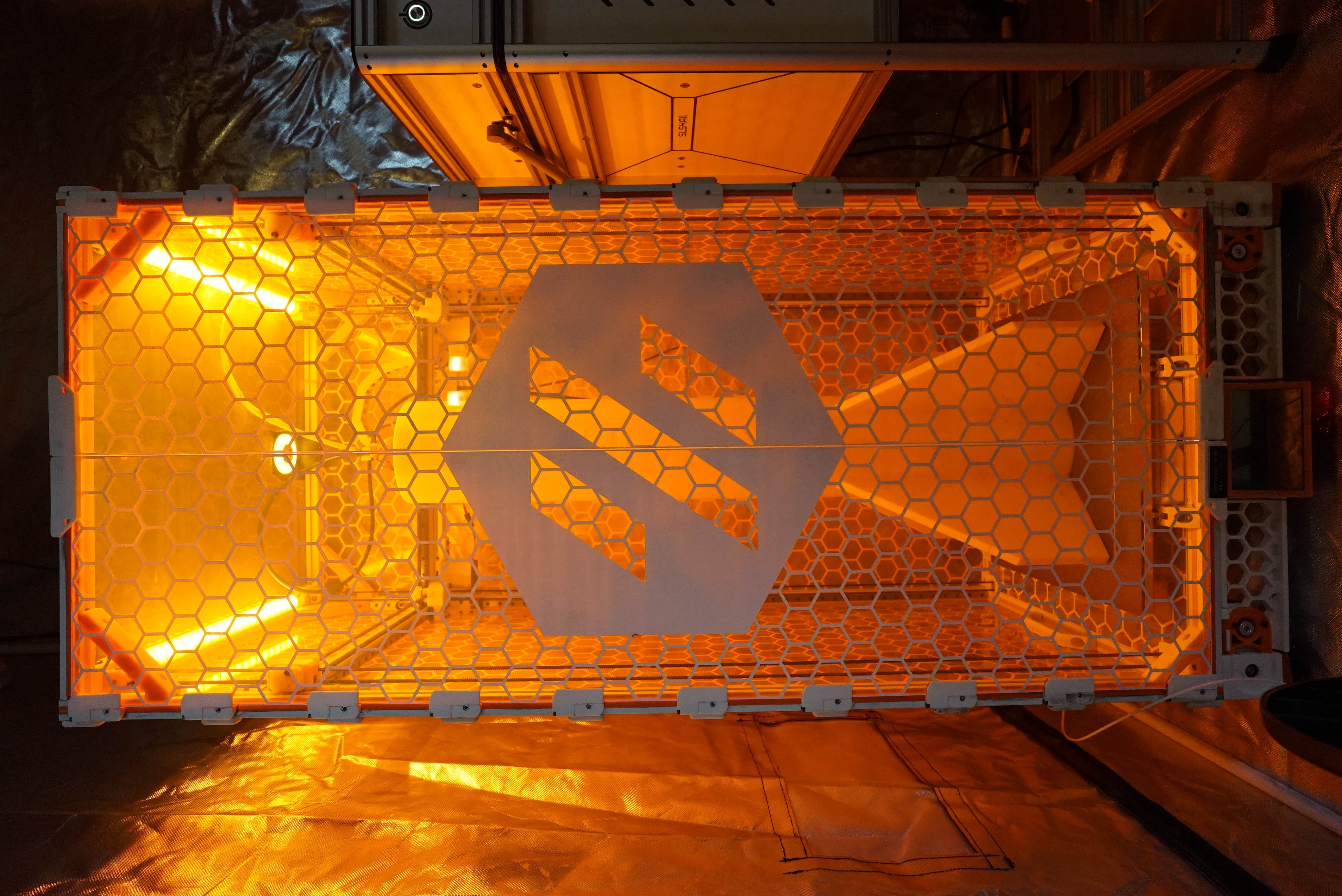}
    \end{minipage}
    \caption{
        (Left) \texttt{Voron-2-Tall} during assembly, upside down to access
        internal electronics underneath the build plate. (Middle) First
        successful \textit{Benchy} print with fully assembled printer and
        \textit{StealthBurner}
        \cite{noauthor_vorondesignvoron-stealthburner_2026} toolhead using PLA
        filament. (Right) \texttt{Voron-2-Tall} with finished print for lower
        airframe component of \textit{Big Heavy 2}
        \cite{pak_ppak10big-heavy_2026} using ABS filament.
    }
    \label{fig:voron_2_tall}
\end{figure}
\clearpage

\section{Onboard Camera Data}
\label{sec:onboard_camera_data}

\begin{figure}
    \centering
    \includegraphics[width=\linewidth]{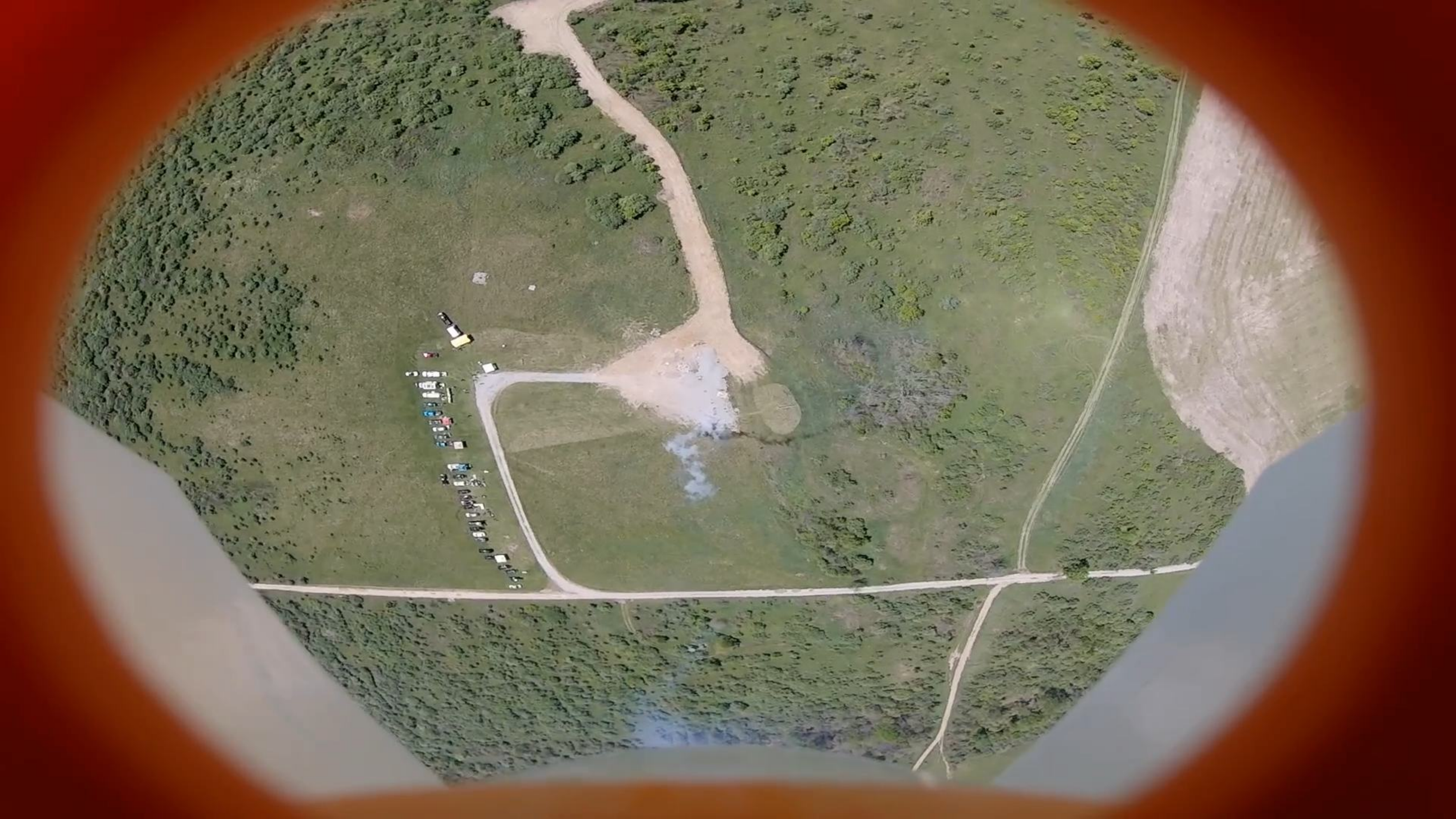}
    \caption{
        View of the launch site taken from High Power 1 onboard RunCam 5 camera
        just before reaching an apogee of 338 m (1108 ft).
    }
    \label{fig:runcam_v38}
\end{figure}

\begin{figure}
    \centering
    \includegraphics[width=\linewidth]{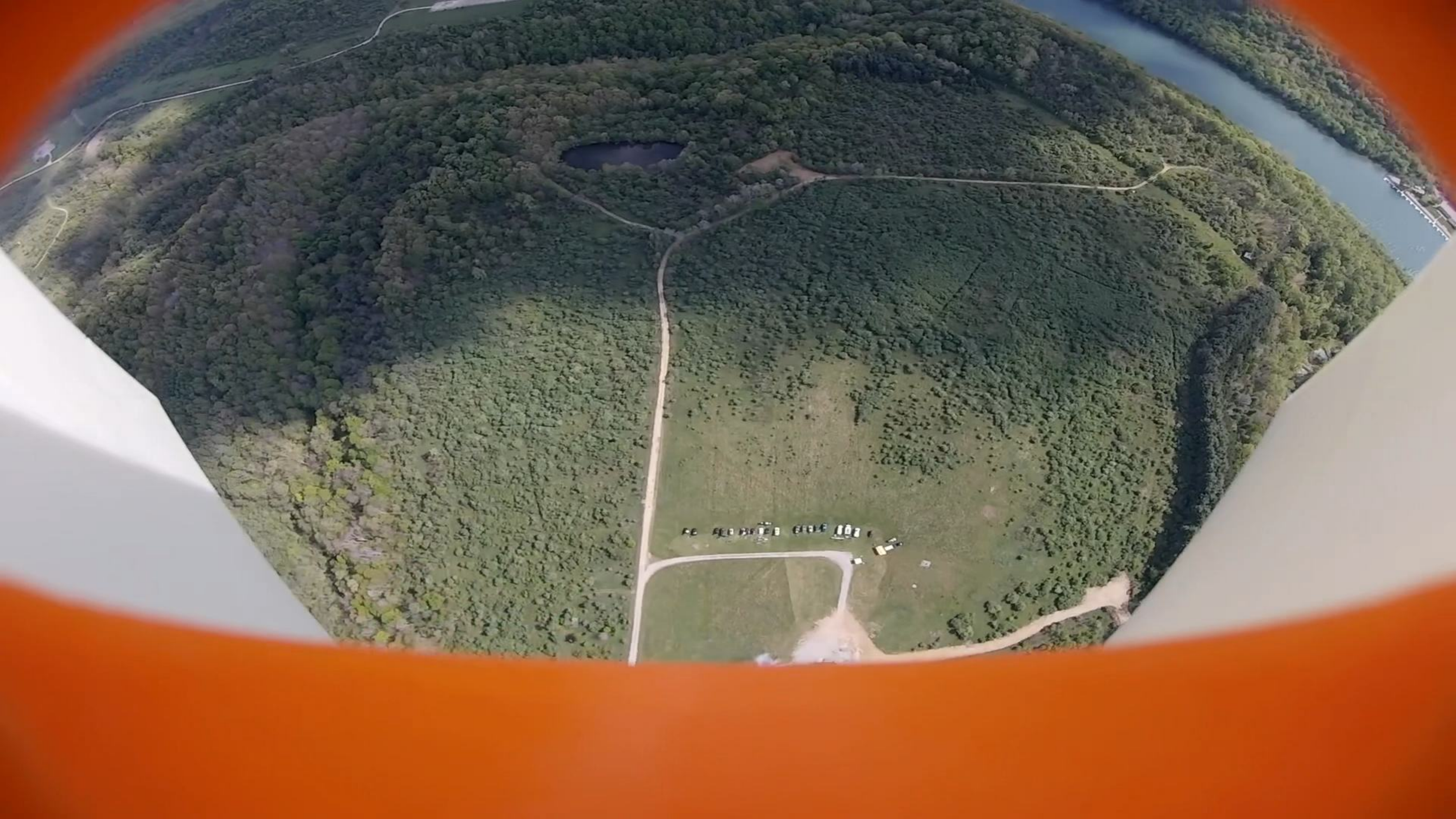}
    \caption{
        Photo taken using RunCam 5 from High Power 4 around measured apogee of
        570 m (1870 ft).
    }
    \label{fig:runcam_big_heavy_2}
\end{figure}
\clearpage

\section{Launch Site Weather Conditions}
\label{sec:launch_site_weather_conditions}

\begin{table}[ht]
\centering
\caption{
    Hourly weather data during Sunday May 3rd, 2026 launch window at Fayette
    County, PA.
}
\label{tab:weather}
\begin{tabular}{rrrrrrr}
\hline
 & & & & \multicolumn{3}{c}{\textbf{Wind}} \\
\cline{5-7}
\textbf{Time} & \textbf{Temperature} & \textbf{Cloud} & \textbf{Humidity} & \textbf{Speed} & \textbf{Direction} & \textbf{Gust} \\
\hline
9 AM  & 41$^\circ$F & 18\% & 72\% & 5.8 mph  & WSW & 11.2 mph \\
10 AM & 45$^\circ$F & 19\% & 60\% & 7.8 mph  & W   & 17.0 mph \\
11 AM & 48$^\circ$F & 76\% & 49\% & 9.6 mph  & W   & 18.7 mph \\
12 PM & 51$^\circ$F & 29\% & 43\% & 10.3 mph & W   & 19.6 mph \\
1 PM  & 53$^\circ$F & 18\% & 39\% & 10.5 mph & W   & 20.3 mph \\
2 PM  & 54$^\circ$F & 14\% & 37\% & 9.6 mph  & W   & 20.6 mph \\
3 PM  & 55$^\circ$F & 13\% & 37\% & 9.6 mph  & WNW & 19.9 mph \\
4 PM  & 56$^\circ$F & 15\% & 38\% & 9.6 mph  & W   & 20.2 mph \\
5 PM  & 56$^\circ$F & 50\% & 40\% & 8.9 mph  & W   & 19.9 mph \\
\hline
\end{tabular}
\end{table}
\clearpage

\bibliography{references}

\end{document}